\documentclass[sigconf]{acmart}
\AtBeginDocument{%
  }

\copyrightyear{2025} 
\acmYear{2025} 
\setcopyright{acmlicensed}\acmConference[KDD '25]{Proceedings of the 31st ACM
SIGKDD Conference on Knowledge Discovery and Data Mining V.1}{August 3--7,
2025}{Toronto, ON, Canada}
\acmBooktitle{Proceedings of the 31st ACM SIGKDD Conference on Knowledge
Discovery and Data Mining V.1 (KDD '25), August 3--7, 2025, Toronto, ON, Canada}
\acmDOI{10.1145/3690624.3709392}
\acmISBN{979-8-4007-1245-6/25/08}

\usepackage{enumitem}
\usepackage{booktabs}
\usepackage{colortbl}
\usepackage{subfig}
\usepackage{multirow}
\usepackage[utf8]{inputenc}
\usepackage[linesnumbered,ruled,vlined]{algorithm2e}
\usepackage{soul, color, xcolor}

\let\titleold\title
\renewcommand{\title}[1]{\titleold{#1}\newcommand{\thetitle}{#1}}
\def\maketitlesupplementary
   {
   \newpage
       \twocolumn[
        \centering
        \Huge
        \vspace{1.5em}\textbf{Supplementary Material} \\
        \vspace{2.5em}
       ] 
   }

\begin{document}

\title{Improving Synthetic Image Detection Towards Generalization: An Image Transformation Perspective}

\author{Ouxiang Li}
\authornote{This work was done during the internship in Xiaohongshu Inc..}
\affiliation{%
  \institution{University of Science and Technology of China}
  \city{Hefei}
  \country{China}
}
\email{lioox@mail.ustc.edu.cn}

\author{Jiayin Cai}
\affiliation{%
  \institution{Xiaohongshu Inc.}
  \city{Beijing}
  \country{China}}
\email{jiayin1@xiaohongshu.com}

\author{Yanbin Hao}
\authornote{Corresponding authors.}
\affiliation{%
  \institution{University of Science and Technology of China}
  \city{Hefei}
  \country{China}}
\email{haoyanbin@hotmail.com}

\author{Xiaolong Jiang}
\affiliation{%
  \institution{Xiaohongshu Inc.}
  \city{Beijing}
  \country{China}}
\email{laige@xiaohongshu.com}

\author{Yao Hu}
\affiliation{%
  \institution{Xiaohongshu Inc.}
  \city{Beijing}
  \country{China}}
\email{xiahou@xiaohongshu.com}

\author{Fuli Feng}
\authornotemark[2]
\affiliation{%
  \institution{University of Science and Technology of China}
  \city{Hefei}
  \country{China}}
\email{fulifeng93@gmail.com}

\renewcommand{\shortauthors}{Ouxiang Li et al.}

\begin{abstract}
With recent generative models facilitating photo-realistic image synthesis, the proliferation of synthetic images has also engendered certain negative impacts on social platforms, thereby raising an urgent imperative to develop effective detectors. Current synthetic image detection (SID) pipelines are primarily dedicated to crafting universal artifact features, accompanied by an oversight about SID training paradigm. In this paper, we re-examine the SID problem and identify two prevalent biases in current training paradigms, \textit{i.e.}, weakened artifact features and overfitted artifact features. Meanwhile, we discover that the imaging mechanism of synthetic images contributes to heightened local correlations among pixels, suggesting that detectors should be equipped with local awareness. In this light, we propose SAFE, a lightweight and effective detector with three simple image transformations. Firstly, for weakened artifact features, we substitute the down-sampling operator with the crop operator in image pre-processing to help circumvent artifact distortion. Secondly, for overfitted artifact features, we include ColorJitter and RandomRotation as additional data augmentations, to help alleviate irrelevant biases from color discrepancies and semantic differences in limited training samples. Thirdly, for local awareness, we propose a patch-based random masking strategy tailored for SID, forcing the detector to focus on local regions at training. Comparative experiments are conducted on an open-world dataset, comprising synthetic images generated by \textbf{26 distinct generative models.} Our pipeline achieves a new state-of-the-art performance, with remarkable improvements of 4.5\% in accuracy and 2.9\% in average precision against existing methods. Our code is available at: \url{https://github.com/Ouxiang-Li/SAFE}.
\end{abstract}




\maketitle

\section{Introduction} \label{sec:intro}
\begin{figure*}[t]
    \centering
    \includegraphics[width=\hsize]{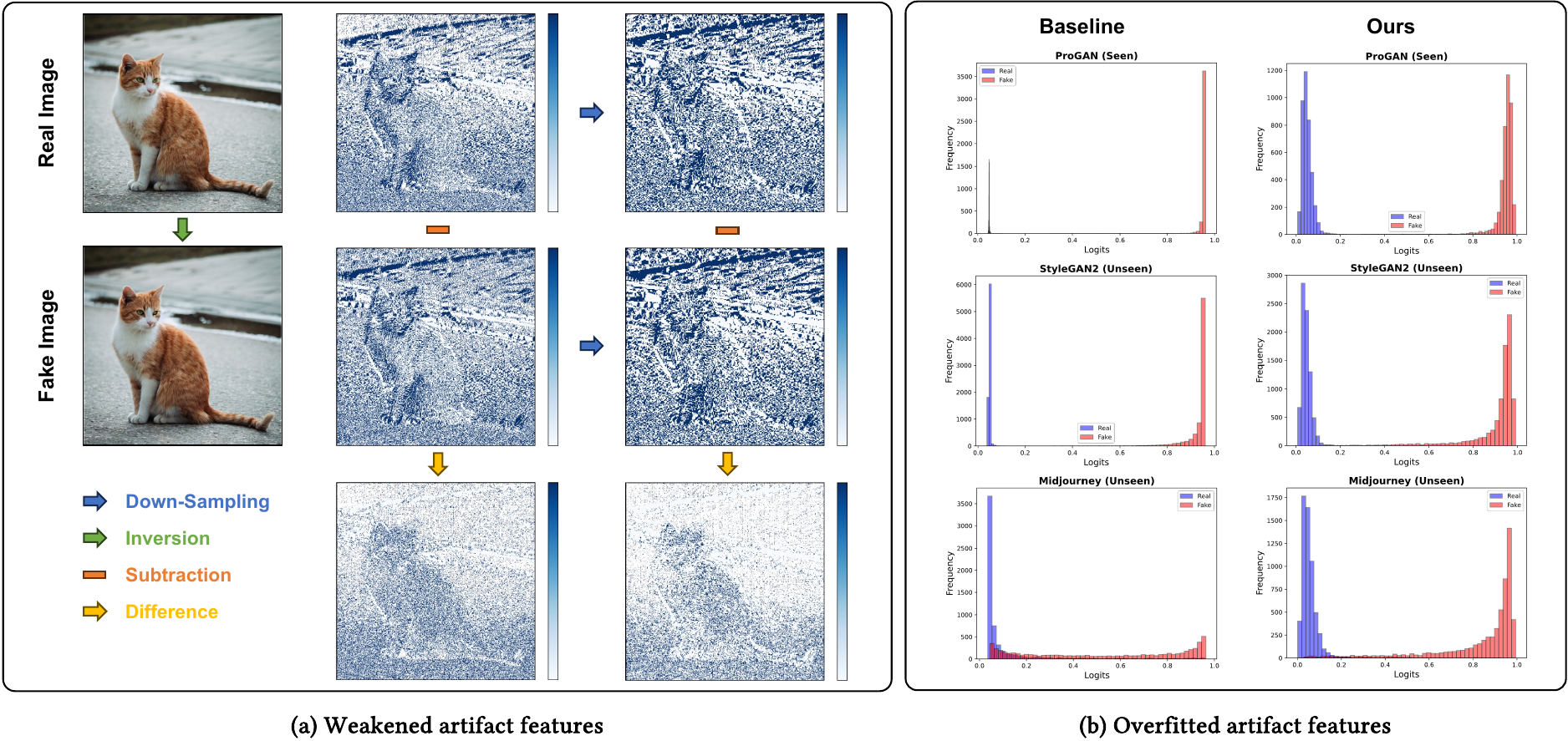}
    \vspace{-6mm}
    \caption{Two prevalent biases observed in current SID training paradigms. (a) Weakened artifact features: We reconstruct the real image using null-text inversion \cite{mokady2022nulltextinversioneditingreal} with Stable Diffusion v1.4 \cite{StableDiffusion}, ensuring they are semantically consistent. We then calculate their local correlation maps\protect\footnotemark\ before and after down-sampling and subtract them for explicit comparison. It can be noticed that fake image exhibits stronger local correlations and the down-sampling operator indeed weakens such subtle artifacts. (b) Overfitted artifact features: We compare the logit distributions between the baseline (w/ HorizontalFlip only, left) and ours (right) for both seen and unseen generators. The monotonous application of HorizontalFlip is insufficient to alleviate overfitting to training samples, resulting in an extreme logit distribution for in-domain samples (\textit{i.e.}, ProGAN) and inferior generalization for out-of-domain samples (\textit{e.g.}, Midjourney).}
    \label{fig:bias}
\end{figure*}

Recent advancements in generative models have substantially simplified the synthesis of photo-realistic images, such as Generative Adversarial Networks (GANs) \cite{goodfellow2014generative} and diffusion models (DMs) \cite{ho2020denoising}. While these technologies are thriving in low-cost image synthesis with unprecedented realism, they simultaneously pose potential societal risks, including disinformation \cite{bontridder2021role}, deepfakes \cite{rossler2019faceforensics++}, and privacy concerns \cite{golda2024privacy,li2024model}. In this context, numerous Synthetic Image Detection (SID) pipelines have been developed, aiming to distinguish between natural (real) and synthetic (fake) images.

SID aims to identify synthetic images from various generative models, termed ``generalization performance''. As novel generative models are persistently developed and deployed, ensuring generalization performance to new generators is particularly critical in the industry. Accordingly, current SID pipelines are dedicated to exploring universal differences between natural and synthetic images, termed ``universal artifact features'', to improve their generalization performance, which can be categorized into two branches. The first image-aware branch leverages image inputs to autonomously extract universal artifact features via detection-aware modules, operating from multiple perspectives, \textit{e.g.}, frequency \cite{qian2020thinking,li2021frequency,liu2024forgery}, semantics \cite{liu2020global,ju2022fusing,koutlis2024leveraging}, and image-text alignment \cite{sha2023fake,keita2024bi}. The second feature-aware branch simplifies the detection pipeline into two stages, with stage-1 manually extracting universal artifact features by means of off-the-shelf models \cite{liu2022detecting,tan2023learning,wang2023dire,ojha2023towards,luo2024lare} or image processing operators \cite{jeong2022bihpf,zhong2023rich,yan2024sanity,tan2024frequency,tan2024rethinking} in advance and stage-2 training the classifier with these pre-processed features. 

\footnotetext{The detailed algorithm is presented in Appx.~\ref{sec:details_of_algorithm}.}
In addition to crafting universal artifact features, we argue that generalizable SID also necessitates reasonable training strategies. For instance, current pipelines typically adopt the down-sampling operator during image pre-processing to standardize images into a uniform shape. However, this common operator is not suitable for SID, which could potentially distort the subtle artifacts, leading to \textbf{weakened artifact features} (\textit{cf.} Fig.~\ref{fig:bias} (a)). Meanwhile, these pipelines also confront \textbf{overfitted artifact features} (\textit{cf.} Fig.~\ref{fig:bias} (b)) due to the monotonous data augmentation (\textit{e.g.}, HorizontalFlip), which is insufficient to bridge the distributional disparity between training and testing data. Afflicted by these feature-agnostic biases, such biased training paradigms can inherently prohibit current pipelines from achieving superior generalization.

In this light, we attribute the inferior generalization of existing detectors to their oversight of SID-specific image transformations at training. Hence, we re-examine the SID problem and identify the following critical factors in effective detection: 

\begin{itemize}[leftmargin=*]
    \item \textbf{Artifact preservation:} By analyzing the imaging mechanisms of synthetic images, we find the widespread use of up-sampling and convolution operators in generative models contributes to heightened local correlations in synthetic images. Consequently, the down-sampling operator can inevitably weaken such local correlations via re-weighting adjacent pixel values.
    
    \item \textbf{Invariant augmentation:} Aiming to generalize to unseen synthetic images with limited training data, the detector has to learn the specified artifact mode without being affected by other irrelevant features. Therefore, artifact-invariant data augmentations will improve the performance of SID detectors.
    
    \item \textbf{Local awareness:} Considering the inherent difference in imaging between natural and synthetic images, local awareness can facilitate the detector to focus more on local correlations in adjacent pixels, which offers a crucial clue for generalizable detection.
\end{itemize}

Building upon above insights, we propose SAFE (\textbf{S}imple Preserved and \textbf{A}ugmented \textbf{FE}atures), a simple and lightweight SID pipeline that integrates three effective image transformations with simple artifact features. 
\textbf{First, w.r.t artifact preservation,} we substitute the conventional down-sampling operator in image pre-processing with the crop operator at both training and inference, which helps circumvent artifact distortion.
\textbf{Second, w.r.t invariant augmentation,} we introduce ColorJitter and RandomRotation as additional data augmentations, which are effective in alleviating irrelevant biases regarding color mode and semantics against limited training samples. 
\textbf{Third, w.r.t local awareness,} we propose a novel patch-based random masking strategy tailored for SID in data augmentation, forcing the detector to focus on local regions at training. 
Moreover, w.r.t artifact selection, considering the existing detectable differences between natural and synthetic images in high-frequency components \cite{ricker2022towards}, we simply introduce Discrete Wavelet Transform (DWT) \cite{mallat1989theory} to extract high-frequency features.
To comprehensively evaluate the generalization performance, we benchmark our proposed pipeline across 26 generators\footnote{ProGAN, StyleGAN, StyleGAN2, BigGAN, CycleGAN, StarGAN, GauGAN, Deepfake, AttGAN, BEGAN, CramerGAN, InfoMaxGAN, MMDGAN, RelGAN, S3GAN, SNGAN, STGAN, DALLE, Glide, ADM, LDM, Midjourney, SDv1.4, SDv1.5, Wukong, VQDM.} with only access to ProGAN \cite{karras2017progressive} generations and corresponding real images at training. Extensive experiments demonstrate the effectiveness of our image transformations in SID even with simple artifact features. 

Our contributions can be summarized as follows:
\begin{itemize}[leftmargin=*]
    \item We attribute the inferior generalization in current detectors to their biased training paradigms without comprehensive analyses of SID-specific image transformations. This observation prompts us to re-examine the SID problem and identify crucial factors that contribute to simple yet effective detection.

    \item We propose SAFE, a simple and lightweight SID pipeline that integrates three effective image transformations with simple artifact features. These techniques are effective in generalizable SID by alleviating training biases and enhancing local awareness.

    \item Extensive experiments demonstrate the effectiveness of our proposed pipeline, exhibiting remarkable generalization performance across 26 generators, with improvements of 4.5\% in accuracy and 2.9\% in average precision compared against state-of-the-art (SOTA) baselines.
\end{itemize}

\section{Related Works} \label{sec:related_work}

\begin{figure}[t]
    \centering
    \includegraphics[width=\hsize]{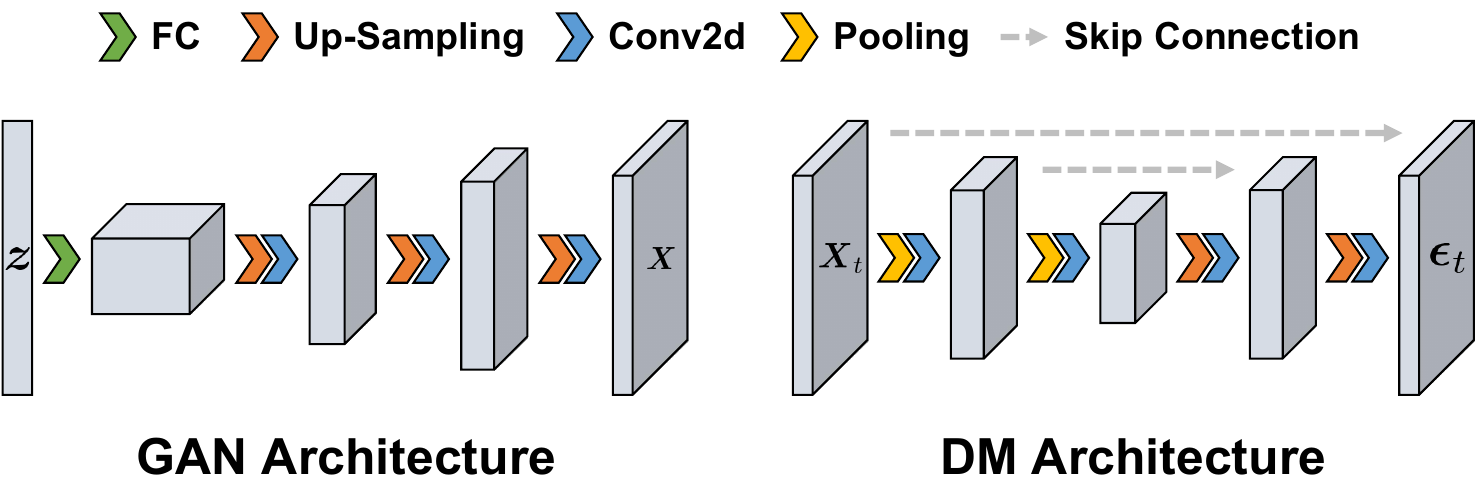}
    \vspace{-6mm}
    \caption{The synthesis mechanisms of two common generative architectures ( \textit{i.e.}, GANs and DMs), where up-sampling and convolution operators are both widely used.}
    \label{fig:architecture}
\end{figure}

In this section, we comprehensively summarize current SID pipelines, categorizing them into two branches as introduced in Sec.~\ref{sec:intro}, \textit{i.e.},  image-aware branch and feature-aware branch.

\subsection{Image-aware Branch}
The image-aware branch aims to autonomously extract universal artifact features with meticulously crafted detection-aware modules from diverse perspectives. From the frequency perspective, F3Net \cite{qian2020thinking} introduces FAD, LFS, and MixBlock for feature extraction and interaction. Later, FDFL \cite{li2021frequency} and FatFormer \cite{liu2024forgery} propose AFFGM and FAA blocks for adaptive frequency forgery extraction, respectively. From the semantic perspective, GramNet \cite{liu2020global} inserts Gram Blocks into different semantic levels from ResNet \cite{he2016deep} to extract global image texture features. Fusing \cite{ju2022fusing} combines global and local embeddings, with PSM extracting local embeddings and AFFM fusing features. RINE \cite{koutlis2024leveraging} leverages the image representations extracted by intermediate Transformer blocks from CLIP-ViT \cite{dosovitskiy2020image,radford2021learning} and employs a TIE module to map them into learnable forgery-aware vectors. From the text-image alignment perspective, DE-FAKE \cite{sha2023fake} introduces CLIP text encoder \cite{vaswani2017attention} as an additional cue for detection on text-to-image models. Bi-LORA \cite{keita2024bi} leverages BLIP2 \cite{li2023blip} combined with LoRA \cite{hu2021lora} tuning techniques to enhance the detection performance. 

\subsection{Feature-aware Branch}
The feature-aware branch simplifies the detection pipeline into two stages with pre-processed features. One principal paradigm introduces off-the-shelf models, attempting to extract artifacts through their pretrained knowledge. LNP \cite{liu2022detecting} extracts learned noise patterns using pretrained denoising networks \cite{zamir2020cycleisp}. LGrad \cite{tan2023learning} employs pretrained CNN models as transform functions to convert images into gradients and leverages these gradients to present universal artifacts. UniFD \cite{ojha2023towards} directly utilizes CLIP features for classification from the semantic perspective. DIRE \cite{wang2023dire} proposes to inverse images into Gaussian noise via DDIM inversion \cite{song2020denoising} and reconstruct the noise into images using pretrained ADM \cite{dhariwal2021diffusion}, where the differences between source and reconstructed images are termed as DIRE for detection. Similarly, \cite{wang2024trace,cazenavette2024fakeinversion,luo2024lare} follow the same reconstruction perspective with pretrained VAEs \cite{kingma2013auto} and DMs. To meet real-time requirements, the other paradigm introduces instant image processing operators, exhibiting superior promise in practical applications. FreDect \cite{frank2020leveraging} is the first pipeline to extract frequency artifacts via Discrete Cosine Transform (DCT) in detecting GAN generations. Subsequently, BiHPF \cite{jeong2022bihpf} adopts bilateral high-pass filters to amplify the effect of frequency-level artifacts. More recently, PatchCraft \cite{zhong2023rich} compares the rich-texture and poor-texture regions via pre-defined filter operators. FreqNet \cite{tan2024frequency} extracts high-frequency representations via Fast Fourier Transform (FFT) and its inverse transform. NPR \cite{tan2024rethinking} analyzes the commonality of CNN-based generators and proposes to capture the generalizable structural artifacts stemming from up-sampling operations in image synthesis.

\section{Method}

In this section, we will elaborate on the details of our SAFE pipeline to demonstrate how simple image transformations can improve the generalization performance in SID.

\subsection{Problem Formulation}
The primary objective of SID is to design a universal classifier for generalizable synthetic image detection. In real-world scenarios, the detector is required to distinguish test samples from $n$ unknown plausible sources. That is
\begin{equation}
    \mathcal{D}_{\text{test}} = \{\mathcal{S}_1, \mathcal{S}_2, ..., \mathcal{S}_n\},  \mathcal{S}_i = \{\boldsymbol{X}_j^i, y_j^i\}_{j=1}^{N_i},
\end{equation}
where $N_i$ is the number of images in the $i^{\text{th}}$ source $\mathcal{S}_i$ comprising both natural ($y_j^i = 0$) and synthetic ($y_j^i = 1$) images from a specific generative model. To achieve this goal, one can resort to training a neural network $\mathcal{M}(\mathcal{I}; \boldsymbol{\theta})$ parameterized by $\boldsymbol{\theta}$ with constrained training data from $\mathcal{D}_{\text{train}}$, which typically includes synthetic images generated by limited generators. Because generators are continuously iterating, it is impossible to exhaustively include them in practical training. 
Herein, according to different detection branches, the model input $\mathcal{I}$ could be either the source images $\boldsymbol{X} \in \{\mathcal{D}_{\text{train}}, \mathcal{D}_{\text{test}}\}$ or the pre-processed artifact features $\boldsymbol{f}(\boldsymbol{X})$ with self-crafted processor $\boldsymbol{f}$.

\begin{figure}[t]
    \centering
    \includegraphics[width=\hsize]{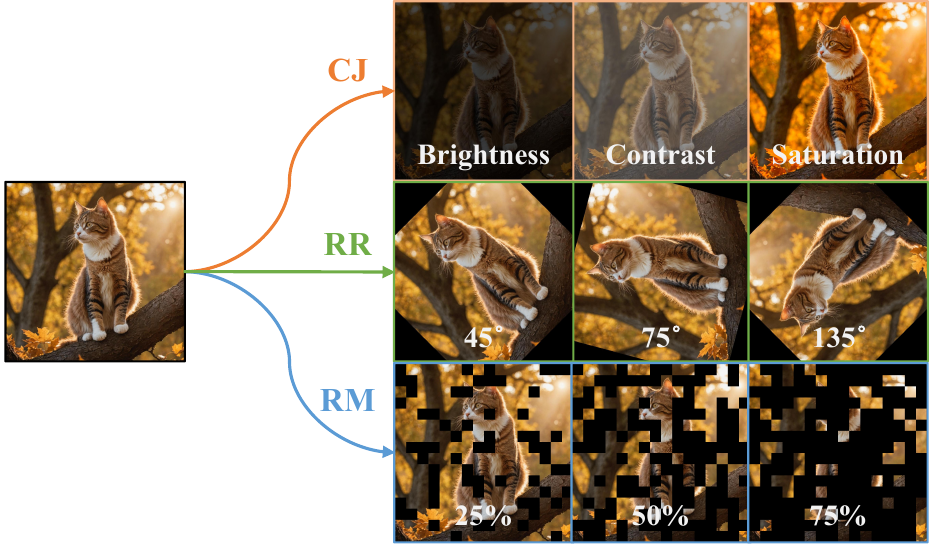}
    \vspace{-6mm}
    \caption{Examples of our proposed transformations in data augmentation, \textit{i.e.}, ColorJitter (CJ), RandomRotatioin (RR), and RandomMask (RM). In practice, these three augmentations are applied simultaneously along with HorizontalFlip.}
    \label{fig:aug}
\end{figure}

\subsection{Artifact Preservation} \label{sec:data_pre_processing}
We first analyze the sources of synthetic artifacts from the perspective of imaging mechanisms for both GANs and DMs in Fig.~\ref{fig:architecture}. In the GAN pipeline, latent features $\boldsymbol{z} \in \mathcal{N}(\mathbf{0}, \mathbf{I})$ are initially transformed from low-resolutions into high-resolutions through a fully-connected layer. Subsequently, up-sampling (\textit{up}) and convolution (\textit{conv}) operators are simultaneously utilized to proceed with image synthesis. Similarly, in the DM pipeline with UNet architecture\footnote{Further discussion on DiT architecture \cite{peebles2023scalable} is presented in Sec.~\ref{sec:latest_generators}.} \cite{ronneberger2015u}, noised images $\boldsymbol{X}_{t}$ at timestep $t$ are first downscaled through pooling and \textit{conv} operators into low-dimensions and then upscaled through \textit{up} and \textit{conv} operators until predicting the noise $\boldsymbol{\epsilon}_{t}$. It can be observed that synthetic images inevitably undergo both \textit{up} and \textit{conv} operators during image synthesis. In numerical computation, both operators can be regarded as a weighted average of pixel values within a neighborhood. This limited receptive field would naturally enhance the \textbf{local correlations} among adjacent pixels in synthetic images, inevitably leaving discriminative artifact features for generalizable detection. 

However, we notice that current SID pipelines typically adopt the down-sampling operator with Bilinear interpolation in image pre-processing to standardize images into a uniform shape. Although this straightforward operation is prevalent in conventional classification tasks, it can inadvertently smooth the noticeable local correlations in synthetic images and thereby weaken those subtle discriminative artifacts in low-level space (see Fig.~\ref{fig:bias} (a)). To tackle this problem, we propose to replace the down-sampling operator with the crop operator, with RandomCrop at training and CenterCrop at inference. This straightforward adjustment facilitates to retain intricate details and subtle local correlations in synthetic images, thereby improving the detector's ability to capture nuanced artifacts from input samples.

\subsection{Artifact Augmentation}
With the refinement of artifact preservation, we observe a noticeable improvement in \textbf{intra-architecture scenarios,} where training and testing images are derived from the same generative architecture, \textit{e.g.}, $\mathcal{D}_{\text{train}} \to \mathcal{D}_{\text{test}}$: ProGAN $\to$ GANs. However, the detector still suffers from inferior generalization performance in \textbf{cross-architecture scenarios,} \textit{e.g.}, $\mathcal{D}_{\text{train}} \to \mathcal{D}_{\text{test}}$: ProGAN $\to$ DMs (see Fig.~\ref{fig:bias} (b)). Given the inherent differences in architectures between GANs and DMs, we attribute this to the detector potentially overfitting certain GAN-specific features. In light of this assertion, we delve into the data augmentation setups and find current pipelines monotonously adopt HorizontalFlip at training. We argue that this extent of data augmentation is insufficient to bridge such distributional disparity in architectures and propose the following techniques to facilitate artifact-invariant augmentation and improve local awareness as shown in Fig.~\ref{fig:aug}.

\textbf{ColorJitter.} The training data is constrained by the limited categories (\textit{e.g.}, car and cat) and generative models, resulting in natural distributional discrepancies between training and testing samples in color mode \cite{li2020identification,xu2024color}. To enhance cross-architecture generalization, we incorporate ColorJitter into data augmentation, adjusting the training distribution by jittering images in color space. The factor controlling how much to jitter images is uniformly sampled from $[\max(0, 1 - \alpha), 1 + \alpha]$, where $\alpha \in [0, 1]$ specifies the allowed perturbation for brightness, contrast, and saturation channels.

\textbf{RandomRotation.} Considering that the local correlations among pixels are robust to the rotation operator, we introduce random rotation as an additional data augmentation. This encourages the detector to focus more on local correlations across different rotation angles, rather than on irrelevant rotation-related features (\textit{e.g.}, semantics). Meanwhile, this simple operation can also enhance the robustness against image rotation. In implementation, we uniformly sample the rotating angle from $[-\beta, +\beta]$, where $\beta \in [0^{\circ}, 180^{\circ}]$, and fill the area outside the rotated image with zero values.

\textbf{RandomMask.} To improve the local awareness of the detector, we propose a random masking technique applied to training samples with a certain probability $p$. In this process, we set the masking patch size to $d \times d$ and the maximum mask ratio to $R \in (0, 1)$. Given an input image of size $H \times W$, we first uniformly sample the actual mask ratio $r \in [0, R]$. Next, we calculate the required number of masking patches using $n = \lfloor (H \times W \times r) / d^2 \rfloor$. These $n$ patches filled with zero-value pixels are then randomly applied to the image, ensuring that there is no overlap among them. We find that masking a high proportion of the input images (\textit{e.g.}, 75\%) still achieves accurate detection during training, demonstrating the detector is capable of distinguishing images based on the remaining unmasked regions. 

\begin{table*}[t]
\centering
\caption{Intra-architecture evaluation on GAN-based synthetic images from ForenSynths \cite{wang2020cnn}. Models here are all trained on \textit{4-class ProGAN}, except for \dag\ trained on the whole training set from ForenSynths, namely \textit{20-class ProGAN}. We report the results in the formulation of ACC (\%) / AP (\%) and average them into ACC$_\text{M}$ (\%) / AP$_\text{M}$ (\%) in the last column. The best result and the second-best result are marked in \textbf{bold} and \underline{underline}, respectively.}
\vspace{-3mm}
\label{table:forensynths}
\resizebox{\hsize}{!}{
\tiny  
\renewcommand{\arraystretch}{1.20}  

\begin{tabular}{l|c|cccccccc|c}
\toprule
Method        & Ref        & ProGAN        & StyleGAN    & StyleGAN2    & BigGAN       & CycleGAN     & StarGAN       & GauGAN       & Deepfake    & Mean        \\ \midrule
CNNDect       & CVPR 2020  & 91.4 / 99.4   & 63.8 / 91.4 & 76.4 / 97.5  & 52.9 / 73.3  & 72.7 / 88.6  & 63.8 / 90.8   & 63.9 / 92.2  & 51.7 / 62.3 & 67.1 / 86.9 \\
FreDect       & ICML 2020  & 90.3 / 85.2   & 74.5 / 72.0 & 73.1 / 71.4  & 88.7 / 86.0  & 75.5 / 71.2  & 99.5 / 99.5   & 69.2 / 77.4  & 60.7 / 49.1 & 78.9 / 76.5 \\
F3Net         & ECCV 2020  & 99.4 / 100.0  & 92.6 / 99.7 & 88.0 / 99.8  & 65.3 / 69.9  & 76.4 / 84.3  & 100.0 / 100.0 & 58.1 / 56.7  & 63.5 / 78.8 & 80.4 / 86.2 \\
BiHPF         & WACV 2022  & 90.7 / 86.2   & 76.9 / 75.1 & 76.2 / 74.7  & 84.9 / 81.7  & 81.9 / 78.9  & 94.4 / 94.4   & 69.5 / 78.1  & 54.4 / 54.6 & 78.6 / 78.0 \\
LGrad         & CVPR 2023  & 99.9 / 100.0  & 94.8 / 99.9 & 96.0 / 99.9  & 82.9 / 90.7  & 85.3 / 94.0  & 99.6 / 100.0  & 72.4 / 79.3  & 58.0 / 67.9 & 86.1 / 91.5 \\
UniFD         & CVPR 2023  & 99.7 / 100.0  & 89.0 / 98.7 & 83.9 / 98.4  & 90.5 / 99.1  & 87.9 / 99.8  & 91.4 / 100.0  & 89.9 / 100.0 & 80.2 / 90.2 & 89.1 / 98.3 \\
PatchCraft\dag    & Arxiv 2023 & 100.0 / 100.0 & 93.0 / 98.9 & 89.7 / 97.8  & 95.7 / 99.3  & 70.0 / 85.1  & 100.0 / 100.0 & 71.9 / 81.8  & 58.6 / 79.6 & 84.9 / 92.8 \\
FreqNet       & AAAI 2024  & 99.6 / 100.0  & 90.2 / 99.7 & 88.0 / 99.5  & 90.5 / 96.0  & 95.8 / 99.6  & 85.7 / 99.8   & 93.4 / 98.6  & 88.9 / 94.4 & 91.5 / 98.5 \\
NPR           & CVPR 2024  & 99.8 / 100.0  & 96.3 / 99.8 & 97.3 / 100.0 & 87.5 / 94.5  & 95.0 / 99.5  & 99.7 / 100.0  & 86.6 / 88.8  & 77.4 / 86.2 & 92.5 / 96.1 \\
FatFormer     & CVPR 2024  & 99.9 / 100.0  & 97.2 / 99.8 & 98.8 / 100.0 & 99.5 / 100.0 & 99.3 / 100.0 & 99.8 / 100.0  & 99.4 / 100.0 & 93.2 / 98.0 & \textbf{98.4} / \textbf{99.7} \\ \rowcolor[HTML]{dadada}
\textbf{Ours} & -          & 99.9 / 100.0  & 98.0 / 99.9 & 98.6 / 100.0 & 89.7 / 95.9  & 98.9 / 99.8  & 99.9 / 100.0  & 91.5 / 97.2  & 93.1 / 97.5 & \underline{96.2} / \underline{98.8} \\ \bottomrule
\end{tabular}

}
\end{table*}
\begin{table*}[t]
\centering
\caption{Intra-architecture evaluation on GAN-based synthetic images from Self-Synthesis \cite{tan2024rethinking}.}
\vspace{-3mm}
\label{table:selfgan}
\resizebox{\hsize}{!}{
\tiny  
\renewcommand{\arraystretch}{1.20}  
\setlength\tabcolsep{4pt}  

\begin{tabular}{l|c|ccccccccc|c}
\toprule
Method        & Ref        & AttGAN       & BEGAN        & CramerGAN    & InfoMaxGAN   & MMDGAN       & RelGAN        & S3GAN        & SNGAN        & STGAN        & Mean        \\ \midrule
CNNDect       & CVPR 2020  & 51.1 / 83.7  & 50.2 / 44.9  & 81.5 / 97.5  & 71.1 / 94.7  & 72.9 / 94.4  & 53.3 / 82.1   & 55.2 / 66.1  & 62.7 / 90.4  & 63.0 / 92.7  & 62.3 / 82.9 \\
FreDect       & ICML 2020  & 65.0 / 74.4  & 39.4 / 39.9  & 31.0 / 36.0  & 41.1 / 41.0  & 38.4 / 40.5  & 69.2 / 96.2   & 69.7 / 81.9  & 48.4 / 47.9  & 25.4 / 34.0  & 47.5 / 54.6 \\
F3Net         & ECCV 2020  & 85.2 / 94.8  & 87.1 / 97.5  & 89.5 / 99.8  & 67.1 / 83.1  & 73.7 / 99.6  & 98.8 / 100.0  & 65.4 / 70.0  & 51.6 / 93.6  & 60.3 / 99.9  & 75.4 / 93.1 \\
LGrad         & CVPR 2023  & 68.6 / 93.8  & 69.9 / 89.2  & 50.3 / 54.0  & 71.1 / 82.0  & 57.5 / 67.3  & 89.1 / 99.1   & 78.5 / 86.0  & 78.0 / 87.4  & 54.8 / 68.0  & 68.6 / 80.8 \\
UniFD         & CVPR 2023  & 78.5 / 98.3  & 72.0 / 98.9  & 77.6 / 99.8  & 77.6 / 98.9  & 77.6 / 99.7  & 78.2 / 98.7   & 85.2 / 98.1  & 77.6 / 98.7  & 74.2 / 97.8  & 77.6 / 98.8 \\
PatchCraft\dag    & Arxiv 2023 & 99.7 / 99.99 & 61.1 / 86.0  & 72.4 / 78.4  & 87.5 / 93.3  & 79.8 / 84.3  & 99.5 / 99.9   & 94.0 / 97.9  & 85.1 / 93.2  & 68.6 / 91.3  & 83.1 / 91.6 \\
FreqNet       & AAAI 2024  & 89.8 / 98.8  & 98.8 / 100.0 & 95.2 / 98.2  & 94.5 / 97.3  & 95.2 / 98.2  & 100.0 / 100.0 & 88.3 / 94.3  & 85.4 / 90.5  & 98.8 / 100.0 & 94.0 / 97.5 \\
NPR           & CVPR 2024  & 83.0 / 96.2  & 99.0 / 99.8  & 98.7 / 99.0  & 94.5 / 98.3  & 98.6 / 99.0  & 99.6 / 100.0  & 79.0 / 80.0  & 88.8 / 97.4  & 98.0 / 100.0 & 93.2 / 96.6 \\
FatFormer     & CVPR 2024  & 99.3 / 99.9  & 99.8 / 100.0 & 98.3 / 100.0 & 98.3 / 99.9  & 98.3 / 100.0 & 99.4 / 100.0  & 99.0 / 99.9  & 98.3 / 99.9  & 98.7 / 99.7  & \underline{98.8} / \textbf{99.9} \\ \rowcolor[HTML]{dadada}
\textbf{Ours} & -          & 99.4 / 100.0 & 99.8 / 100.0 & 99.7 / 100.0 & 99.6 / 100.0 & 99.7 / 100.0 & 99.6 / 100.0  & 94.5 / 100.0 & 98.8 / 100.0 & 99.9 / 100.0 & \textbf{99.0} / \underline{99.8} \\ \bottomrule
\end{tabular}

}
\end{table*}

\subsection{Artifact Selection}
Inspired by recent works on frequency analysis \cite{frank2020leveraging,ricker2022towards,tan2024frequency}, synthetic images are still exhibiting remarkable discrepancy with natural images in high-frequency components. Therefore, we simply introduce DWT as the frequency transform function to extract high-frequency components, which is capable of preserving the spatial structure of images compared with FFT and DCT. Specifically, DWT decomposes the input image $\boldsymbol{X} \in \mathbb{R}^{C \times H \times W}$ into 4 distinct frequency sub-bands $\boldsymbol{X}_{\text{LL}}, \boldsymbol{X}_{\text{LH}}, \boldsymbol{X}_{\text{HL}}, \boldsymbol{X}_{\text{HH}} \in \mathbb{R}^{C \times \frac{H}{2} \times \frac{W}{2}}$, where ``L'' and ``H'' stand for low- and high-pass filters, respectively. Herein, we simply extract the HH-band component $\boldsymbol{X}_{\text{HH}}$ as the input artifact feature for training the detector, which is generalizable enough to distinguish both GAN- and DM-generated images along with our proposed image transformations above.

\section{Experiments}
\subsection{Experimental Setups}

\begin{table*}[t]
\centering
\caption{Cross-architecture evaluation on DM-based synthetic images from Ojha \cite{ojha2023towards}.}
\vspace{-3mm}
\label{table:ojha}
\resizebox{\hsize}{!}{
\tiny  
\renewcommand{\arraystretch}{1.20}  

\begin{tabular}{l|c|cccccccc|c}
\toprule
Method        & Ref        & DALLE       & Glide\_100\_10 & Glide\_100\_27 & Glide\_50\_27 & ADM    & LDM\_100     & LDM\_200     & LDM\_200\_cfg & Mean        \\ \midrule
CNNDect       & CVPR 2020  & 51.8 / 61.3 & 53.3 / 72.9  & 53.0 / 71.3  & 54.2 / 76.0 & 54.9 / 66.6  & 51.9 / 63.7  & 52.0 / 64.5  & 51.6 / 63.1 & 52.8 / 67.4 \\
FreDect       & ICML 2020  & 57.0 / 62.5 & 53.6 / 44.3  & 50.4 / 40.8  & 52.0 / 42.3 & 53.4 / 52.5  & 56.6 / 51.3  & 56.4 / 50.9  & 56.5 / 52.1 & 54.5 / 49.6 \\
F3Net         & ECCV 2020  & 71.6 / 79.9 & 88.3 / 95.4  & 87.0 / 94.5  & 88.5 / 95.4 & 69.2 / 70.8  & 74.1 / 84.0  & 73.4 / 83.3  & 80.7 / 89.1 & 79.1 / 86.5 \\
LGrad         & CVPR 2023  & 88.5 / 97.3 & 89.4 / 94.9  & 87.4 / 93.2  & 90.7 / 95.1 & 86.6 / 100.0 & 94.8 / 99.2  & 94.2 / 99.1  & 95.9 / 99.2 & 90.9 / 97.3 \\
UniFD         & CVPR 2023  & 89.5 / 96.8 & 90.1 / 97.0  & 90.7 / 97.2  & 91.1 / 97.4 & 75.7 / 85.1  & 90.5 / 97.0  & 90.2 / 97.1  & 77.3 / 88.6 & 86.9 / 94.5 \\
PatchCraft\dag    & Arxiv 2023 & 83.3 / 93.0 & 80.1 / 92.0  & 83.4 / 93.9  & 77.6 / 88.7 & 80.9 / 90.5  & 88.9 / 97.7  & 89.3 / 97.9  & 88.1 / 96.9 & 84.0 / 93.8 \\
FreqNet       & AAAI 2024  & 97.2 / 99.7 & 87.8 / 96.0  & 84.4 / 96.6  & 86.6 / 95.8 & 67.2 / 75.4  & 97.8 / 99.9  & 97.4 / 99.9  & 97.2 / 99.9 & 89.5 / 95.4 \\
NPR           & CVPR 2024  & 94.5 / 99.5 & 98.2 / 99.8  & 97.8 / 99.7  & 98.2 / 99.8 & 75.8 / 81.0  & 99.3 / 99.9  & 99.1 / 99.9  & 99.0 / 99.9 & \underline{95.2} / 97.4 \\
FatFormer     & CVPR 2024  & 98.7 / 99.8 & 94.6 / 99.5  & 94.1 / 99.3  & 94.3 / 99.2 & 75.9 / 91.9  & 98.6 / 99.8  & 98.5 / 99.8  & 94.8 / 99.2 & 93.7 / \underline{98.6} \\ \rowcolor[HTML]{dadada}
\textbf{Ours} & -          & 97.5 / 99.7 & 97.3 / 99.4  & 95.8 / 98.9  & 96.6 / 99.2 & 82.4 / 95.8  & 98.8 / 100.0 & 98.8 / 100.0 & 98.7 / 99.9 & \textbf{95.7} / \textbf{99.1} \\ \bottomrule
\end{tabular}

}
\end{table*}
\begin{table*}[t]
\centering
\caption{Cross-architecture evaluation on DM and GAN-based synthetic images from GenImage \cite{zhu2024genimage}.}
\vspace{-3mm}
\label{table:genimage}
\resizebox{\hsize}{!}{
\tiny  
\renewcommand{\arraystretch}{1.20}  

\begin{tabular}{l|c|cccccccc|c}
\toprule
Method        & Ref        & Midjourney  & SDv1.4      & SDv1.5      & ADM         & Glide       & Wukong      & VQDM        & BigGAN      & Mean        \\ \midrule
CNNDect       & CVPR 2020  & 50.1 / 53.4 & 50.2 / 55.8 & 50.3 / 56.3 & 53.0 / 69.2 & 51.7 / 66.9 & 51.4 / 62.4 & 50.1 / 53.6 & 69.7 / 91.8 & 53.3 / 63.7 \\
FreDect       & ICML 2020  & 32.1 / 36.0 & 28.8 / 34.7 & 28.9 / 34.6 & 62.9 / 70.2 & 42.9 / 42.2 & 35.9 / 38.0 & 72.1 / 84.2 & 26.1 / 34.7 & 41.2 / 46.8 \\
LGrad         & CVPR 2023  & 73.7 / 77.6 & 76.3 / 79.1 & 77.1 / 80.3 & 51.9 / 51.4 & 49.9 / 50.5 & 73.2 / 75.6 & 52.8 / 52.0 & 40.6 / 39.3 & 61.9 / 63.2 \\
UniFD         & CVPR 2023  & 57.5 / 69.8 & 65.1 / 81.6 & 64.7 / 81.1 & 69.3 / 84.6 & 60.3 / 74.3 & 73.5 / 88.5 & 86.3 / 95.5 & 89.8 / 97.1 & 70.8 / 84.1 \\
PatchCraft\dag    & Arxiv 2023 & 89.7 / 96.2 & 95.0 / 98.9 & 95.0 / 98.9 & 81.6 / 93.3 & 83.5 / 93.9 & 90.9 / 97.4 & 88.2 / 95.9 & 91.5 / 97.8 & \underline{89.4} / \underline{96.5} \\
FreqNet       & AAAI 2024  & 69.8 / 78.9 & 64.2 / 74.3 & 64.9 / 75.6 & 83.3 / 91.4 & 81.6 / 88.8 & 57.7 / 66.9 & 81.7 / 89.6 & 90.5 / 94.9 & 74.2 / 82.6 \\
NPR           & CVPR 2024  & 77.8 / 85.4 & 78.6 / 84.0 & 78.9 / 84.6 & 69.7 / 74.6 & 78.4 / 85.7 & 76.1 / 80.5 & 78.1 / 81.2 & 80.1 / 88.2 & 77.2 / 83.0 \\
FatFormer     & CVPR 2024  & 56.0 / 62.7 & 67.7 / 81.1 & 68.0 / 81.0 & 78.4 / 91.7 & 87.9 / 95.9 & 73.0 / 85.8 & 86.8 / 96.9 & 96.7 / 99.5 & 76.9 / 86.8 \\ \rowcolor[HTML]{dadada}
\textbf{Ours} & -          & 95.3 / 99.5 & 99.4 / 99.9 & 99.3 / 99.9 & 82.1 / 96.7 & 96.3 / 99.3 & 98.2 / 99.8 & 96.3 / 99.6 & 97.8 / 99.8 & \textbf{95.6} / \textbf{99.3} \\ \bottomrule
\end{tabular}

}
\end{table*}

\noindent \textbf{Training datasets.} Owing to the rapid advancements in generative models, we adhere to the standard protocol derived from ForenSynths \cite{wang2020cnn}, where training data is access to only one generative model and corresponding real images. The training set consists of 20 distinct classes, each comprising 18,000 synthetic images generated by ProGAN \cite{karras2017progressive} and an equal number of real images from the LSUN dataset \cite{yu2015lsun}. In line with previous pipelines \cite{tan2023learning,tan2024rethinking,liu2024forgery}, we adopt the specific 4-class training setting (\textit{i.e., car, cat, chair, horse}), termed as \textit{4-class ProGAN}.

\noindent \textbf{Testing datasets.} To evaluate the generalization performance of different SID pipelines in real-world scenarios, we introduce various natural images from different sources and synthetic images generated by diverse GANs and DMs. Generally, the testing dataset comprises \textbf{4 widely-used datasets} with \textbf{26 generative models}:

\begin{table}[t]
\centering
\caption{We also compare the number of model parameters and FLOPs along with the detection performance (ACC$_\text{M}$ / AP$_\text{M}$) averaged over 33 test subsets from 26 generative models. Extra computational overheads from instant image processing operators (\textit{e.g.}, FFT, DCT, DWT) are too small to be counted in FLOPs.}
\vspace{-3mm}
\label{table:average}
\resizebox{\hsize}{!}{
\tiny  
\renewcommand{\arraystretch}{1.20}  

\begin{tabular}{l|c|ccc}
\toprule
Method        & Ref        & \#Parameters& \#FLOPs  & Mean        \\ \midrule
CNNDect       & CVPR 2020  & 25.56M      & 5.41B   & 59.0 / 75.5 \\
FreDect       & ICML 2020  & 25.56M      & 5.41B   & 55.3 / 56.8 \\
LGrad         & CVPR 2023  & 48.61M      & 50.95B   & 76.6 / 83.1 \\
UniFD         & CVPR 2023  & 427.62M     & 77.83B  & 81.0 / 94.1 \\
PatchCraft\dag& Arxiv 2023 & \textbf{0.12M}       & 6.57B   & 85.3 / 93.6 \\
FreqNet       & AAAI 2024  & 1.85 M      & \underline{3.00B}   & 87.5 / 93.6 \\
NPR           & CVPR 2024  & \underline{1.44M}       & \textbf{2.30B}   & 89.6 / 93.4 \\
FatFormer     & CVPR 2024  & 577.25M     & 127.95B & \underline{92.2} / \underline{96.4} \\ \rowcolor[HTML]{dadada}
\textbf{Ours} & -          & \underline{1.44M}       & \textbf{2.30B}   & \textbf{96.7} / \textbf{99.3} \\ \bottomrule
\end{tabular}

}
\end{table}
\begin{table}[t]
\centering
\caption{Cross-architecture evaluation on DiT-based synthetic images from our DiTFake.}
\vspace{-3mm}
\label{table:ditfake}
\resizebox{\hsize}{!}{
\tiny  
\renewcommand{\arraystretch}{1.20}  

\begin{tabular}{l|ccc|c}
\toprule
Method        & Flux        & PixArt       & SD3         & Mean        \\ \midrule
UniFD         & 52.0 / 62.0 & 53.8 / 67.1  & 53.5 / 66.4 & 53.1 / 65.2 \\
PatchCraft\dag& 90.9 / 97.9 & 94.4 / 99.0  & 93.7 / 98.7 & \underline{93.0} / \underline{98.5} \\
FreqNet       & 73.2 / 80.1 & 59.4 / 65.4  & 66.3 / 73.7 & 66.3 / 73.1 \\
NPR           & 79.5 / 88.7 & 78.5 / 86.0  & 78.9 / 87.5 & 79.0 / 87.4 \\
FatFormer     & 54.4 / 61.8 & 67.1 / 79.1  & 58.3 / 69.1 & 59.9 / 70.0 \\ \rowcolor[HTML]{dadada}
\textbf{Ours} & 99.3 / 99.9 & 99.6 / 100.0 & 99.4 / 99.9 & \textbf{99.4} / \textbf{99.9} \\ \bottomrule
\end{tabular}

}
\end{table}

\begin{itemize}[leftmargin=*]
    \item \textbf{8 models from ForenSynths \cite{wang2020cnn}:} This testset includes real images sampled from 6 datasets (\textit{i.e.}, LSUN \cite{yu2015lsun}, ImageNet \cite{deng2009imagenet}, CelebA \cite{liu2015deep}, CelebA-HQ \cite{karras2018progressive}, COCO \cite{lin2014microsoft}, and FaceForensics++ \cite{rossler2019faceforensics++}) and fake images derived from 8 generators\footnote{ProGAN \cite{karras2017progressive}, StyleGAN \cite{karras2019style}, StyleGAN2 \cite{karras2020analyzing}, BigGAN \cite{brock2018large}, CycleGAN \cite{zhu2017unpaired}, StarGAN \cite{choi2018stargan}, GauGAN \cite{park2019semantic}, Deepfake \cite{rossler2019faceforensics++}.} with the same categories, where Deepfake images are partially forged from real face images.
    
    \item \textbf{9 GANs from Self-Synthesis \cite{tan2024rethinking}:} To further enrich the existing GAN-based test scenes, additional 9 GANs\footnote{AttGAN \cite{AttGAN}, BEGAN \cite{began}, CramerGAN \cite{CramerGAN}, InfoMaxGAN \cite{InfoMaxGAN}, MMDGAN \cite{MMDGAN}, RelGAN \cite{RelGAN}, S3GAN \cite{S3GAN}, SNGAN \cite{SNGAN}, and STGAN \cite{STGAN}.} have been introduced, each generating 4,000 synthetic images. These are accompanied by an equal number of real images, providing a robust dataset for evaluating the generalization performance of SID detectors across a diverse range of GAN variants.
    
    \item \textbf{4 DMs from Ojha \cite{ojha2023towards}:} This testset sources real images from the LAION dataset \cite{schuhmann2021laion} and incorporates 4 recent text-to-image DMs\footnote{DALLE \cite{ramesh2022hierarchical}, Glide \cite{nichol2021glide}, ADM \cite{dhariwal2021diffusion}, LDM \cite{rombach2022high}.} to generate fake images based on text descriptions. For the Glide series, the author introduces Glide generations with varying denoising and up-sampling steps, including Glide\_100\_10, Glide\_100\_27, and Glide\_50\_10.
    In the LDM series, in addition to LDM\_200 which uses 200 denoising steps, generations with classifier-free guidance (LDM\_200\_cfg) and with fewer denoising steps (LDM\_100) are also included.
    
    \item \textbf{7 DMs and 1 GAN from GenImage \cite{zhu2024genimage}:} This testset collects real images of 1,000 classes from the ImageNet dataset \cite{deng2009imagenet} and generates fake images conditioned on the same 1,000 classes with 8 SOTA generators\footnote{Midjourney \cite{midjourney}, SDv1.4 \cite{StableDiffusion}, SDv1.5 \cite{StableDiffusion}, ADM \cite{dhariwal2021diffusion}, Glide \cite{nichol2021glide}, Wukong \cite{wukong}, VQDM \cite{gu2022vector}, BigGAN \cite{brock2018large}.}. Each test subset consists of 6,000 to 8,000 synthetic images and an equivalent number of real images. Additionally, this dataset includes synthetic images of varying dimensions, ranging from $128^2$ to $1024^2$, which poses an additional challenge for existing SID pipelines.
\end{itemize}

\noindent \textbf{Baselines.} To demonstrate that simple transformations can still improve SID performance without complicated designs, we introduce 10 representative baselines for comparison, including CNNDect \cite{wang2020cnn}, FreDect \cite{frank2020leveraging}, F3Net \cite{qian2020thinking}, BiHPF \cite{jeong2022bihpf}, LGrad \cite{tan2023learning}, UniFD \cite{ojha2023towards}, PatchCraft \cite{zhong2023rich}, FreqNet \cite{tan2024frequency}, NPR \cite{tan2024rethinking}, and FatFormer \cite{liu2024forgery}.

\noindent \textbf{Evaluation metrics.} The classification accuracy (ACC) and average precision (AP) are introduced as the main metrics in evaluating the SID performance across various generators. To intuitively evaluate the detection performance on GANs and DMs, we also report the averaged metrics for each testset, termed ACC$_\text{M}$ and AP$_\text{M}$.

\noindent \textbf{Implementation details.} We introduce a lightweight ResNet \cite{he2016deep} from \cite{tan2024rethinking} with only 1.44M parameters to meet real-time requirements. For image pre-processing, we apply random cropping of $256^2$ at training and center cropping of $256^2$ at testing. In terms of data augmentations, we set $\alpha = 0.5$ for ColorJitter, $\beta = 180^{\circ}$ for RandomRotation, and $p=0.5, d = 16, R = 75\%$ for RandomMask. The DWT is configured with symmetric mode and bior1.3 wavelet. The detector is trained using AdamW optimizer \cite{loshchilov2017decoupled} with batch size of 32, learning rate of $5 \times 10^{-3}$, weight decay of 0.01, for 20 epochs on 4 Nvidia H800 GPUs. Besides, the warmup epoch is set to 1 and a cosine annealing scheduler is adopted for the rest epochs.

\subsection{Generalization Comparisons}

\noindent \textbf{Generalization on GAN-based testsets.} We first evaluate the intra-architecture scenario (\textit{i.e.}, ProGAN $\to$ GANs) on two GAN-based testsets in Table~\ref{table:forensynths} and Table~\ref{table:selfgan}. Our SAFE pipeline demonstrates competitive performance against the SOTA method FatFormer with simple image transformations and artifact features. In contrast, FatFormer ensembles CLIP semantics, frequency artifacts, and text-modality guidance together within its pipeline, which brings burdensome computations and latencies in real-world applications. Simultaneously, our results are superior to the other latest pipelines (\textit{e.g.}, FreqNet and NPR). Notably, our detection on Deepfake achieves considerable results with 93.1\% ACC, while most pipelines are generally struggling with this testset since fake images in Deepfake are partially forged from real images. This challenges the detector's ability to differentiate based on local modifications, emphasizing the necessity of local awareness in SID.

\noindent \textbf{Generalization on DM-based testsets.} To further evaluate the generalization performance in the cross-architecture scenario (\textit{i.e.}, ProGAN $\to$ DMs), we introduce another two DM-based testsets in Table~\ref{table:ojha} and Table~\ref{table:genimage}. It can be observed that our SAFE pipeline achieves superior performance against all baselines under this scenario. FatFormer, which achieves SOTA performance in GAN-based detection, exhibits inferior performance in DM-based detection. This disparity suggests that its detection pipeline is more precisely tailored to GAN architectures, rendering it suboptimal for DM-based detection. Additionally, we notice a significant improvement in the GenImage testset, which incorporates the lasted generative models across diverse image categories. Compared with UniFD using semantic features for detection, such improvement indicates that our detector is robust to semantic variations, highlighting the significance of using low-level features rather than semantics. Because iterative advancements in generative models facilitate synthetic images increasingly indistinguishable in semantics, hindering generalizable detection from the semantic perspective.

\begin{table}[t]
\centering
\caption{Daily average recall volume ($\times 10^3$) under different precisions and corresponding improvement (imp.), where the model is deployed online to detect two categories (\textit{i.e.}, human portraits and animals).}
\vspace{-3mm}
\label{table:online}
\small  

\begin{tabular}{cccc}
\toprule
     & P95                    & P98                   & P99        \\ \midrule
Base & 92.7                   & 12.3                  & 9.6        \\
Ours & 135.2                  & 19.2                  & 14.9       \\
\textbf{Imp.} & \textbf{+ 45.85\%}     & \textbf{+ 56.10\%}    & \textbf{+ 55.21\%}  \\ \bottomrule
\end{tabular}

\end{table}

\noindent \textbf{Overall evaulation.} In real-world scenarios, computational complexity and generalization performance are typically the two most crucial metrics for SID. Therefore, we compare all baselines regarding model parameters, FLOPs, and averaged detection results in Table~\ref{table:average}. Our SAFE pipeline reaches superior performance with the least FLOPs, with improvements of 4.5\% in ACC and 2.9\% in AP. In contrast, FatFormer achieves the second-best detection performance in the sacrifice of burdensome computational complexity, with 577.25M parameters and 127.95B FLOPs. Our improvements in both computational complexity and detection performance signify the effectiveness of the proposed image transformations even with simple artifact features in SID. Meanwhile, this also prompts us to rethink the functioning of self-crafted features in existing pipelines, questioning whether they are genuinely more generalizable for SID or merely mitigate potential biases in certain aspects.

\begin{figure}[t]
    \centering
    \subfloat[ACC$_\text{M}$]{\includegraphics[width=.49\columnwidth]{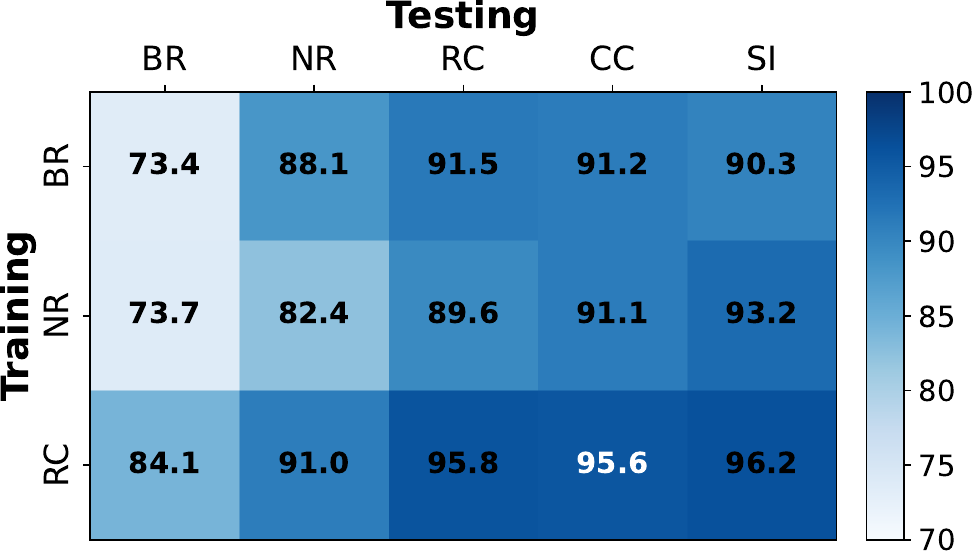}}\hspace{2pt}
    \subfloat[AP$_\text{M}$]{\includegraphics[width=.49\columnwidth]{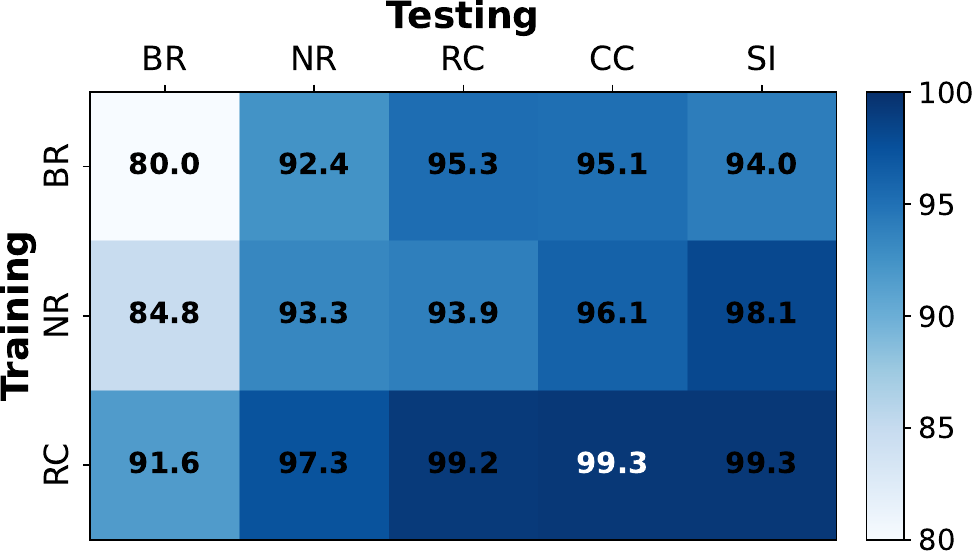}}\\
    \vspace{-3mm}
    \caption{Image pre-processing ablation. We ablate different data pre-processing operators at both training and testing on GenImage. The training includes Bilinear-based Resize (BR), Nearest-based Resize (NR), and RandomCrop (RC). The testing includes BR, NR, RC, CenterCrop (CC), and Source Image (SI), where SI indicates inference w/o any pre-processing. Our pipeline with RC (training) and CC (testing) is marked with white color.}
    \label{fig:pp_abaltion}
\end{figure}
\begin{figure}[t]
    \centering
    \subfloat[ACC$_\text{M}$]{\includegraphics[width=.49\columnwidth]{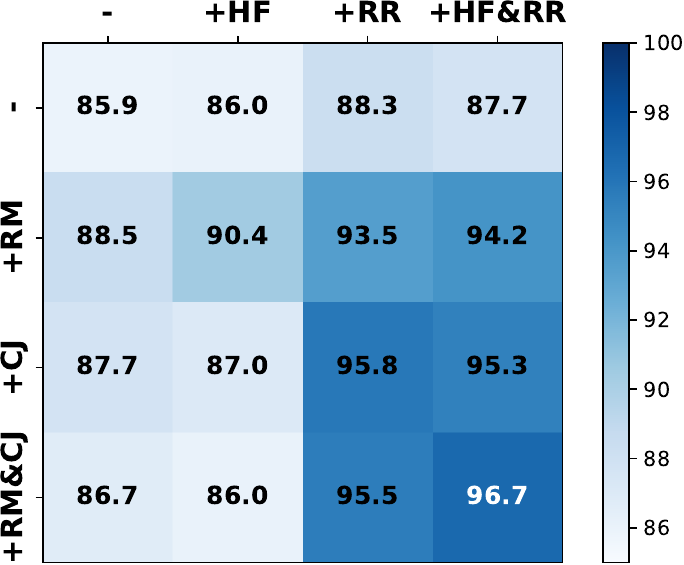}}\hspace{2pt}
    \subfloat[AP$_\text{M}$]{\includegraphics[width=.49\columnwidth]{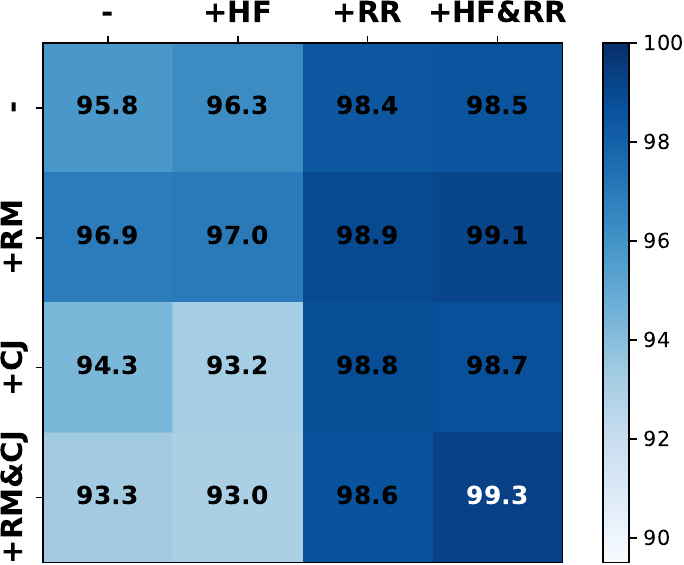}}\\
    \vspace{-3mm}
    \caption{Image augmentation ablation. We ablate the introduced data augmentation techniques, including HorizontalFlip (HF), RandomRotation (RR), RandomMask (RM), and ColorJitter (CJ), where ``\textbf{$+$}" and ``\textbf{$-$}" indicate w/ and w/o a specific augmentation, respectively. Our pipeline combined with all augmentations is marked with white color.}
    \label{fig:da_abaltion}
\end{figure}

\subsection{Generalization to Latest Generators} \label{sec:latest_generators}
We also notice that recent Diffusion Transformers (DiTs) \cite{peebles2023scalable} have successfully advanced the transition of DMs from convolution-based to transformer-based architectures, leading to the emergence of new SOTA generators. To evaluate the generalization of current SID detectors to these latest generators, we propose DiTFake, a testset that comprises generations from three DiT-based models  (\textit{i.e.}, Flux \cite{Flux}, PixArt \cite{chen2024pixart}, and SD3 \cite{esser2024scaling}) and an equal number of real images from COCO \cite{lin2014microsoft}. Further details about this testset can be found in Appx.~\ref{sec:ditfake_appendix}. 
It can be observed that DiTFake is a high-quality testset for SID and poses significant challenges to existing detectors. Even so, our method outperforms all baselines with ACC$_\text{M}$ of 99.4\% from Table~\ref{table:ditfake}, demonstrating our superior generalization against novel generators despite being trained solely on ProGAN. Notable, these DiT-based generators do not include specific up-sampling and convolution operators in their architectures, which seems to contradict our analysis in Sec.~\ref{sec:data_pre_processing}. Nevertheless, their model designs all follow the LDM paradigm \cite{rombach2022high}, \textit{i.e.}, they all sample images in the latent space and then upscale them into image space through a VAE \cite{kingma2013auto}. This VAE is a convolutional model with multiple up-sampling and convolution operators, inevitably leaving artifacts in synthetic images as well. On top of that, our detector can generalize well to these novel generators, validating the effectiveness of our method in learning unbiased and universal artifacts for generalizable SID.

\subsection{Online Experiment}
We conduct an online experiment on our production platform with around 9 million user-generated content per day. Compared with the previous online detector (base) that uses FFT amplitude and phase spectrum, our pipeline achieves a recall volume improvement of around 50\% under the same precision in Table~\ref{table:online}. This comparison demonstrates that our pipeline can achieve more generalizable detection in business data by mitigating training biases and enhancing local awareness with simple image transformations.

\begin{figure}[t]
    \centering
    \subfloat[ACC$_\text{M}$]{\includegraphics[width=.49\columnwidth]{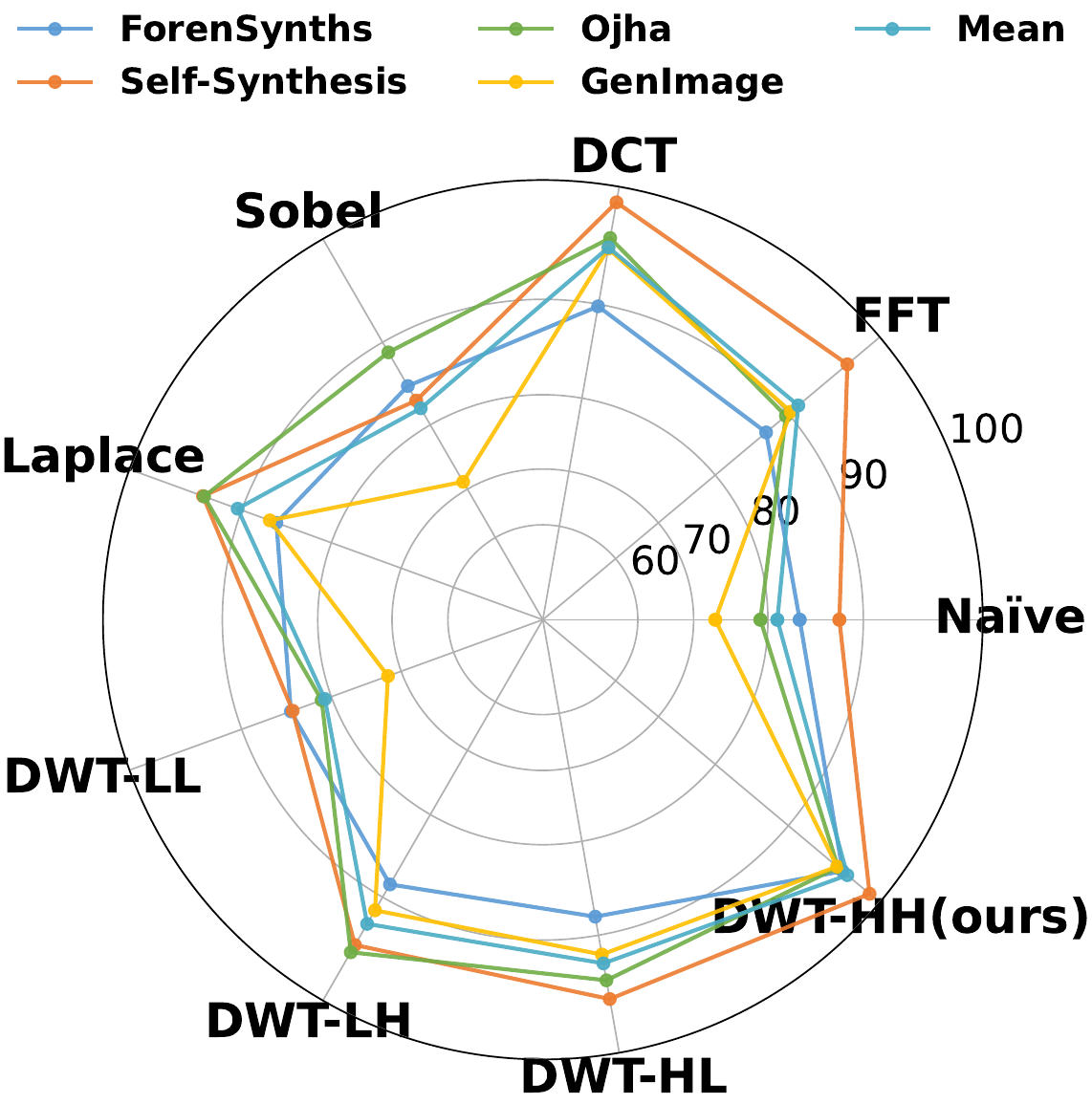}}\hspace{2pt}
    \subfloat[AP$_\text{M}$]{\includegraphics[width=.49\columnwidth]{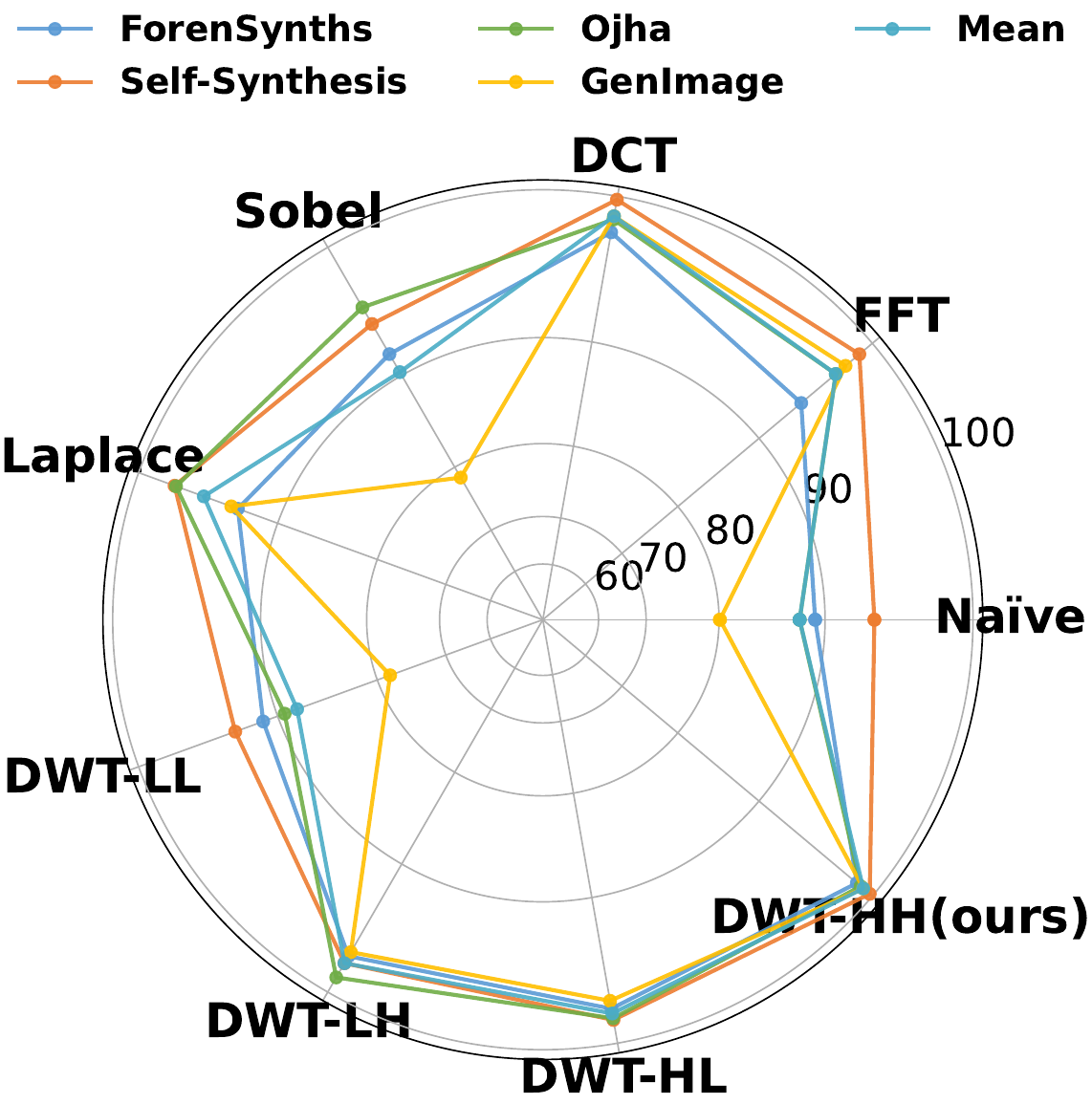}}\\
    \vspace{-3mm}
    \caption{\textbf{Feature selection ablation.} We compare the Naïve baseline (\textit{i.e.}, trained with source images) and different low-level feature extractors mainly adopted in frequency transforms (\textit{i.e.}, FFT, DCT, DWT) and edge detection (\textit{i.e.}, Sobel, Laplace), where LL, LH, HL, HH represent 4 distinct frequency bands in DWT. Details can be found in Appx.~\ref{sec:details_of_table_ll}.
    }
    \label{fig:ll_ablation}
\end{figure} 

\subsection{Ablation Studies} \label{sec:ablation_studies}
To thoroughly comprehend the effects of our proposed image transformations along with selected artifact features in SID, we conduct extensive ablations. 

\noindent \textbf{Image pre-processing.} In real-world scenarios, images inevitably undergo various operations, with resizing being the most common. In Fig.~\ref{fig:pp_abaltion}, we compare the detection performance trained with different image pre-processing operators (\textit{i.e.}, BR, NR, and RC) and report ACC$_\text{M}$ and AP$_\text{M}$ on GenImage, which includes images with various dimensions of $128^2$, $256^2$, $512^2$, and $1024^2$, etc. The operation image size is set to $256^2$ except for SI, which retains the original image size. We can draw three conclusions: (1) Even though the testset undergoes resize operators, our training strategy (RC) still achieves superior performance than BR and NR, which can be attributed to the artifact-preserved training with crop operators. (2) The resize operator indeed diminishes the subtle artifact features and degrades the detection performance, necessitating the crop operator during training. (3) The similar performance between CC, RC, and SI during testing indicates that our pipeline can achieve accurate detection through center-cropped regions only, suggesting 
our detector is translation-robust to cropped regions of test images.

\begin{figure}[t]
    \centering
    \subfloat[ACC]{\includegraphics[width=.49\columnwidth]{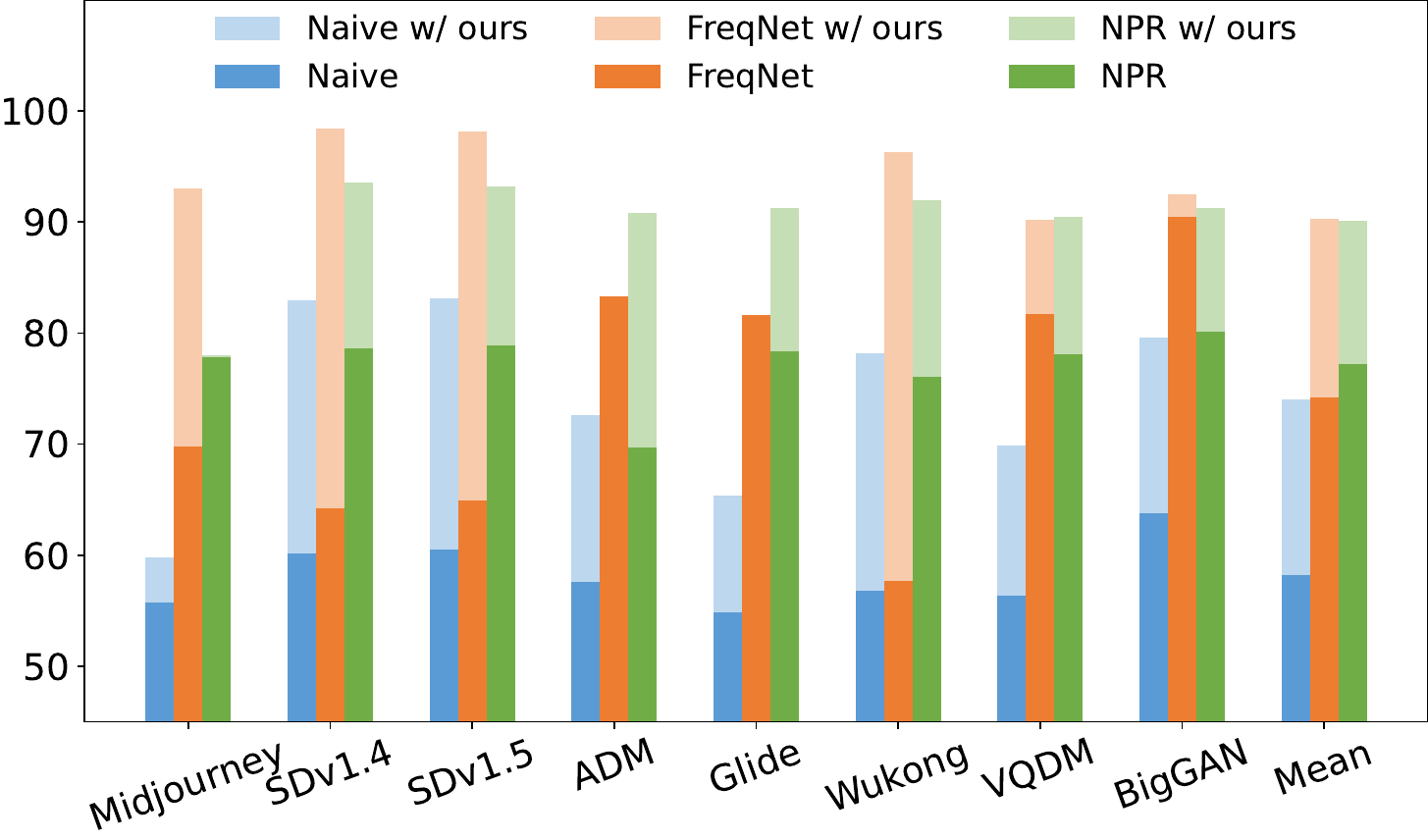}}\hspace{2pt}
    \subfloat[AP]{\includegraphics[width=.49\columnwidth]{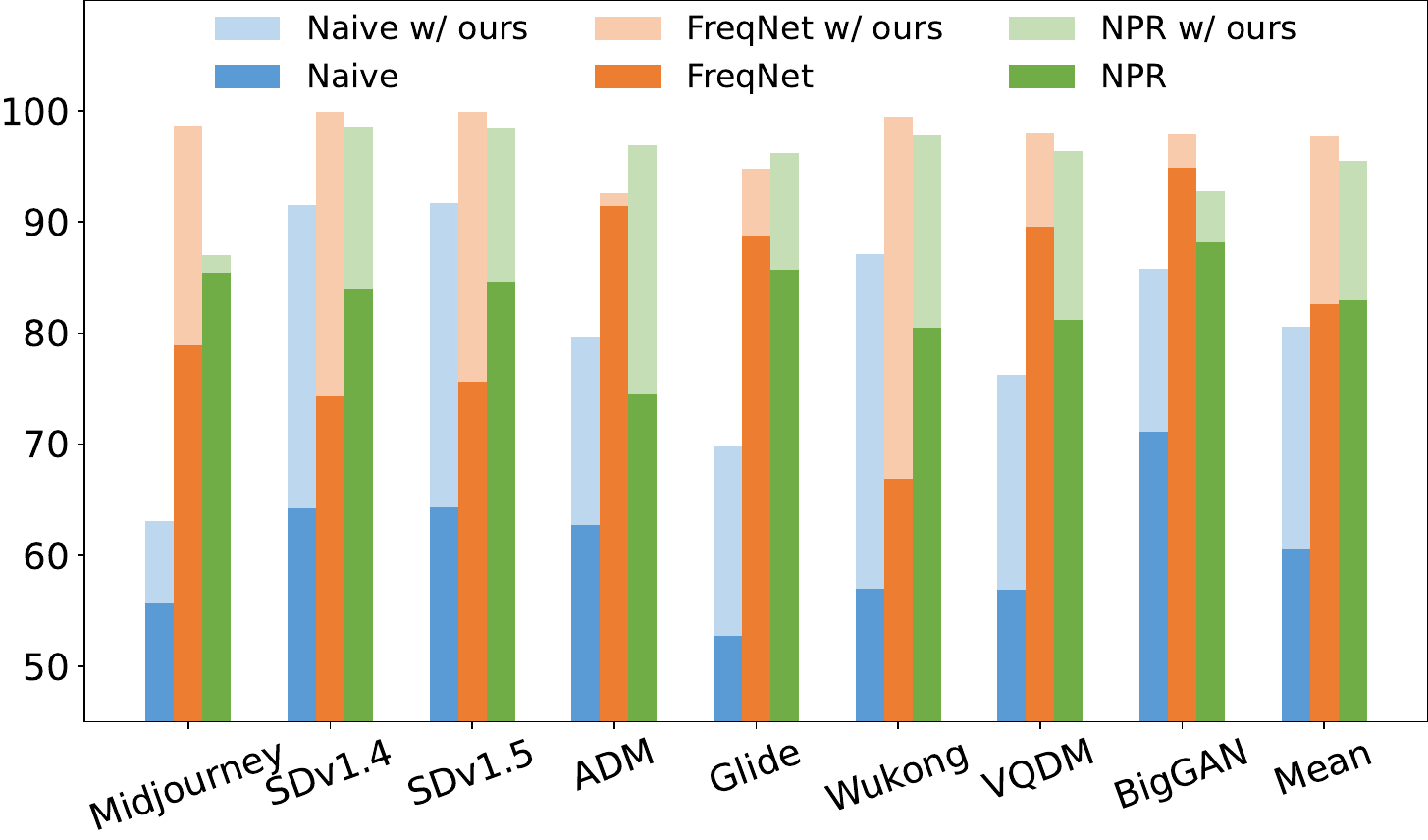}}\\
    \vspace{-3mm}
    \caption{Plug \& Play application with existing pipelines. We compare the Naïve baseline, FreqNet, and NPR on the GenImage testset due to its diverse range of image dimensions, generator types, and data categories.}
    \label{fig:plug_and_play}
\end{figure}

\noindent \textbf{Image augmentation.} We then ablate the proposed data augmentations along with HorizontalFlip (HF) and report the average results on all 33 testsets in Fig.~\ref{fig:da_abaltion}. It can be observed that HF exerts almost no effect on improving generalization performance, which is insufficient to mitigate overfitting in previous pipelines. In contrast, the proposed three techniques (\textit{i.e.}, CJ, RR, and RM) can each independently enhance the generalization performance by augmenting training images, thereby bridging the distributional disparity among synthetic images from different architectures to alleviate overfitting biases and enhance local awareness. Moreover, their benefits to generalization performance are cumulative, achieving optimal results of 96.7\% ACC$_\text{M}$ and 99.3\% AP$_\text{M}$ when adopted in combination.

\noindent \textbf{Feature selection.} We also compare various artifact feature extractors commonly used in digital image processing, as shown in Fig.~\ref{fig:ll_ablation}. Our findings are as follows: (1) Low-level features demonstrate superior generalization compared to the Naïve baseline trained with source images, particularly in cross-architecture scenarios. This intuitive comparison aligns with our motivation for incorporating frequency features. (2) Regarding frequency component selection, high-frequency components exhibit better generalization than low-frequency ones. The inferior performance of generative models in capturing high-frequency details makes these details a useful discriminative clue. (3) For high-frequency extractors, methods such as FFT, DCT, and DWT all show favorable generalization performance, with DWT performing the best. We have integrated DWT into our pipeline because it allows direct extraction of high-frequency features in the spatial domain without additional inverse transformations and manual frequency filtering.

\noindent \textbf{Plug \& Play application.} Since our method is model-agnostic, we apply our proposed image transformations to existing SID pipelines as a plug-and-play module. The comparison results, illustrated in Fig.~\ref{fig:plug_and_play}, demonstrate a consistent improvement in detecting synthetic images from various generative models. This indicates that these image transformations can help the detector learn more preserved and generalizable artifact features, thereby improving its ability to capture nuanced artifacts from input samples with enhanced generalization.

\section{Conclusion}
In this paper, we have re-examined current SID pipelines and discovered that these pipelines are inherently prohibited by biased training paradigms from superior generalization. In this light, we propose a simple yet effective pipeline, SAFE, to alleviate these training biases with three simple image transformations.
Our pipeline integrates crop operators in image pre-processing and combines ColorJitter, RandomRotation, and RandomMask in image augmentation, with the DWT extracting high-frequency artifact features. Extensive experiments demonstrate the effectiveness of our pipeline in both computational efficiency and detection performance even with simple artifacts. This prompts us to rethink the rationale behind current pipelines dedicated to various self-crafted features, wondering whether these features are genuinely more generalizable in SID or merely mitigate potential biases in certain aspects.
We hope our findings will facilitate further endeavors regarding mitigating biases in SID training paradigms before exploring self-crafted artifacts.

\bibliographystyle{ACM-Reference-Format}
\bibliography{sample-base}

\appendix
\clearpage
\maketitlesupplementary

\noindent This Appendix is organized as follows:
\begin{itemize}[leftmargin=*]
    \item In Sec.~\ref{sec:experimental_details_appendix}, we elaborate on more experimental details omitted in our main paper because of page limits.
    \item In Sec.~\ref{sec:additional_experiments_appendix}, we conduct additional experiments about hyperparameter sensitivity and detection robustness for a more comprehensive evaluation.
    \item In Sec.~\ref{sec:ditfake_appendix}, we describe in detail the construction process of the DiTFake dataset and provide additional visualizations.
    \item In Sec.~\ref{sec:limitation_appendix}, we summarize the plausible limitations of our pipelines and expect to address them in future work.
\end{itemize}

\begin{figure*}[t]
    \centering
    \subfloat[Jitter Factor $\alpha$]{\includegraphics[width=.24\hsize]{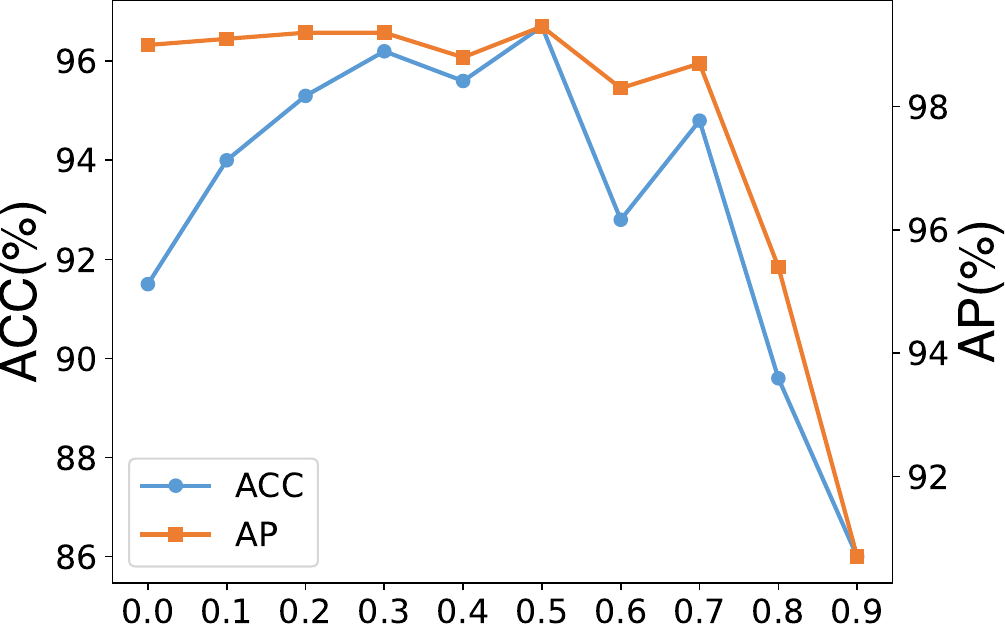}}\hspace{2pt}
    \subfloat[Rotation Angle $\beta$]{\includegraphics[width=.24\hsize]{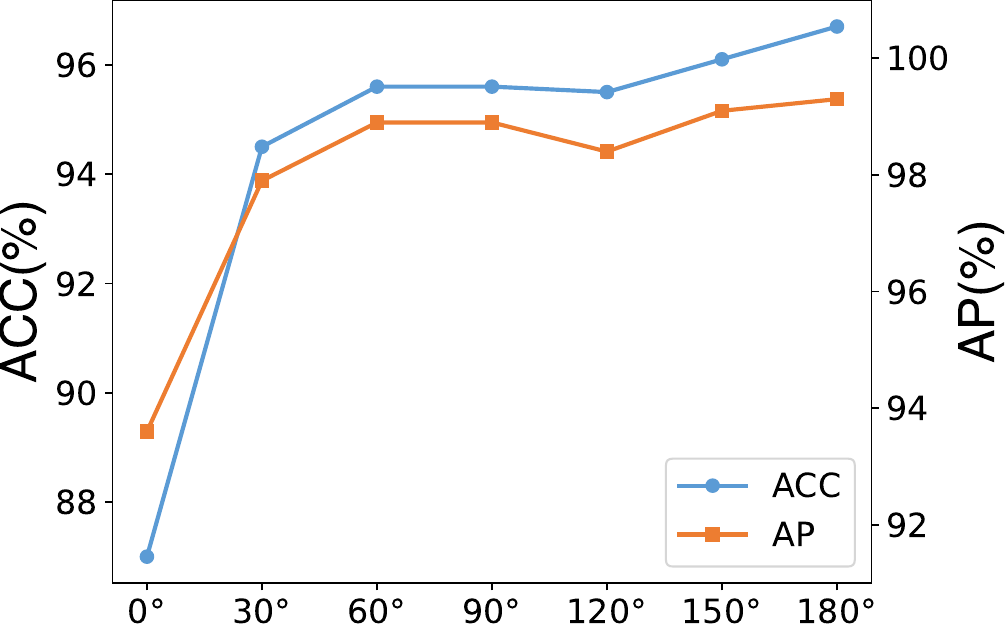}}\hspace{2pt}
    \subfloat[Patch Size $d$]{\includegraphics[width=.24\hsize]{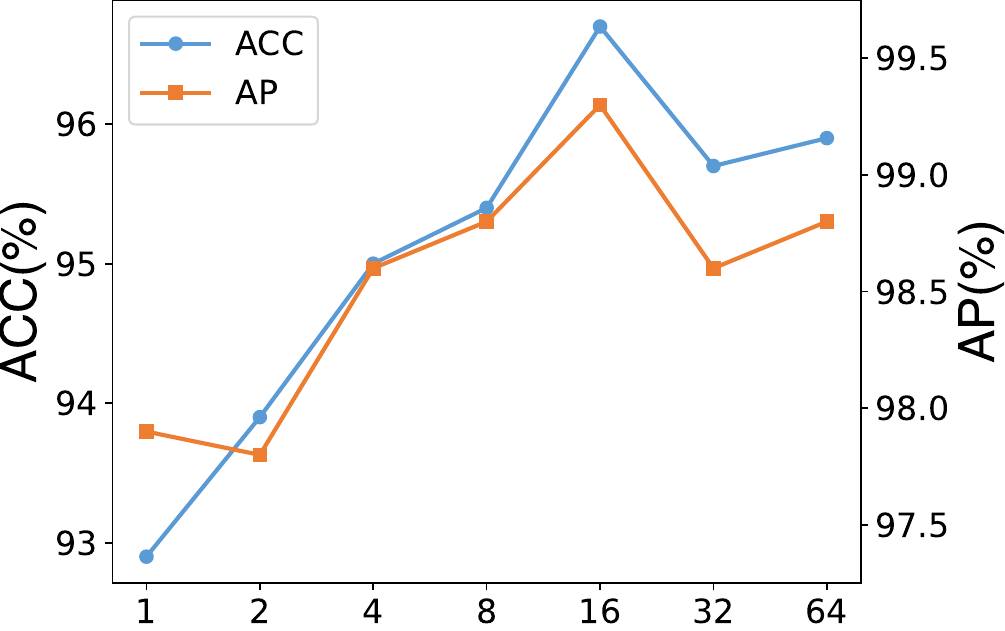}}\hspace{2pt}
    \subfloat[Mask Ratio $R$]{\includegraphics[width=.24\hsize]{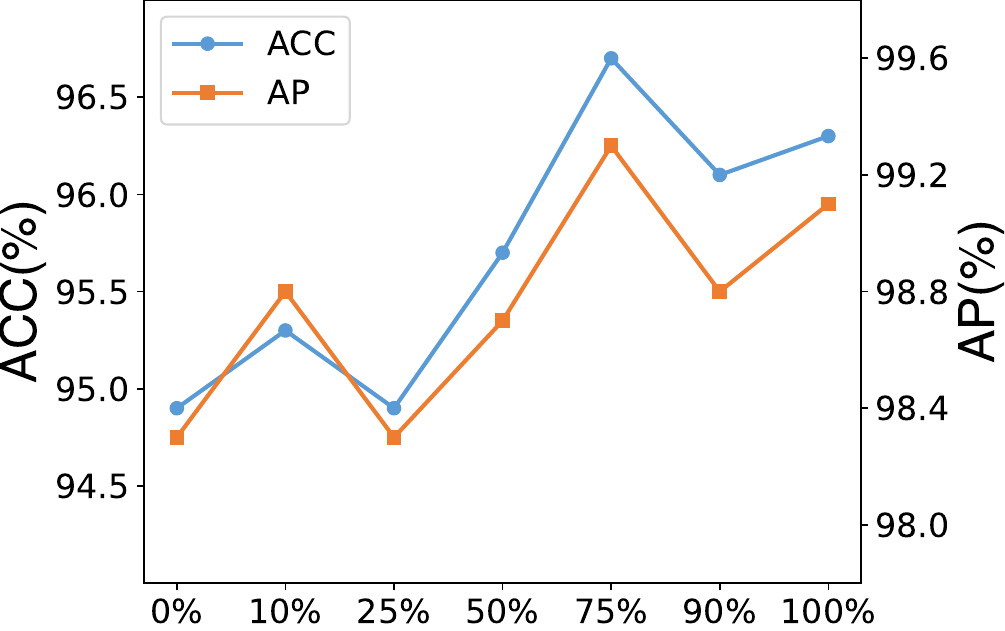}}
    \vspace{-3mm}
    \caption{Ablation study of hyperparameters in artifact augmentations, including jitter factor $\alpha$ in ColorJitter (CJ), rotation angle $\beta$ in RandomRotation (RR), patch size $d$ and mask ratio $R$ in RandomMask (RM).}
    \label{fig:hyper_ablation_appendix}
\end{figure*}
\begin{figure}[t]
    \centering
    \subfloat[ACC$_\text{M}$]{\includegraphics[width=.49\columnwidth]{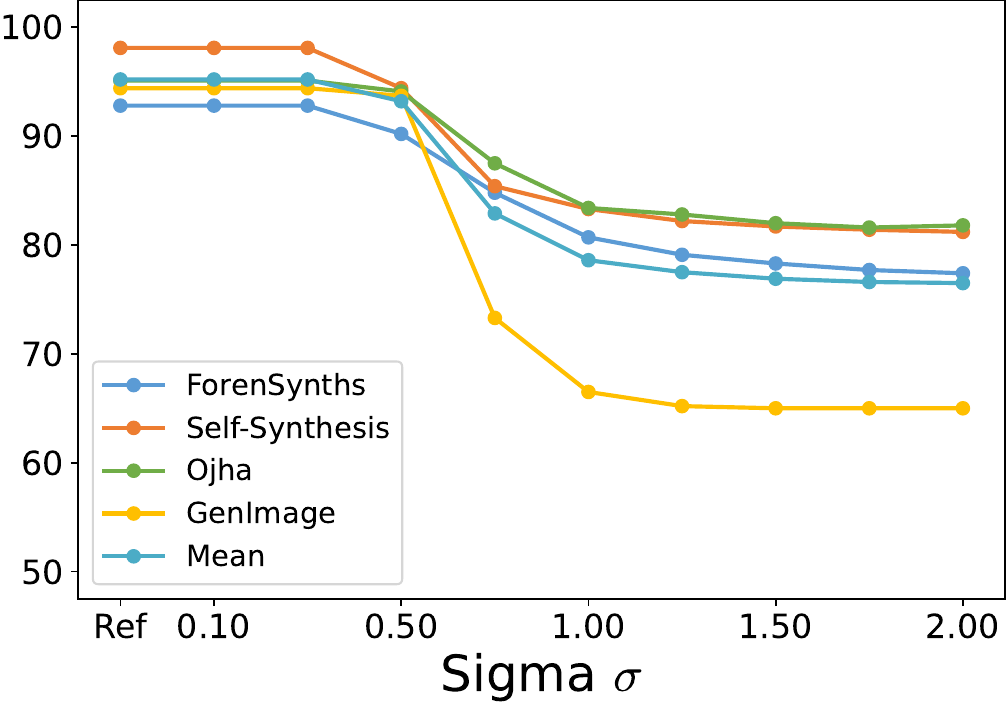}}\hspace{2pt}
    \subfloat[AP$_\text{M}$]{\includegraphics[width=.49\columnwidth]{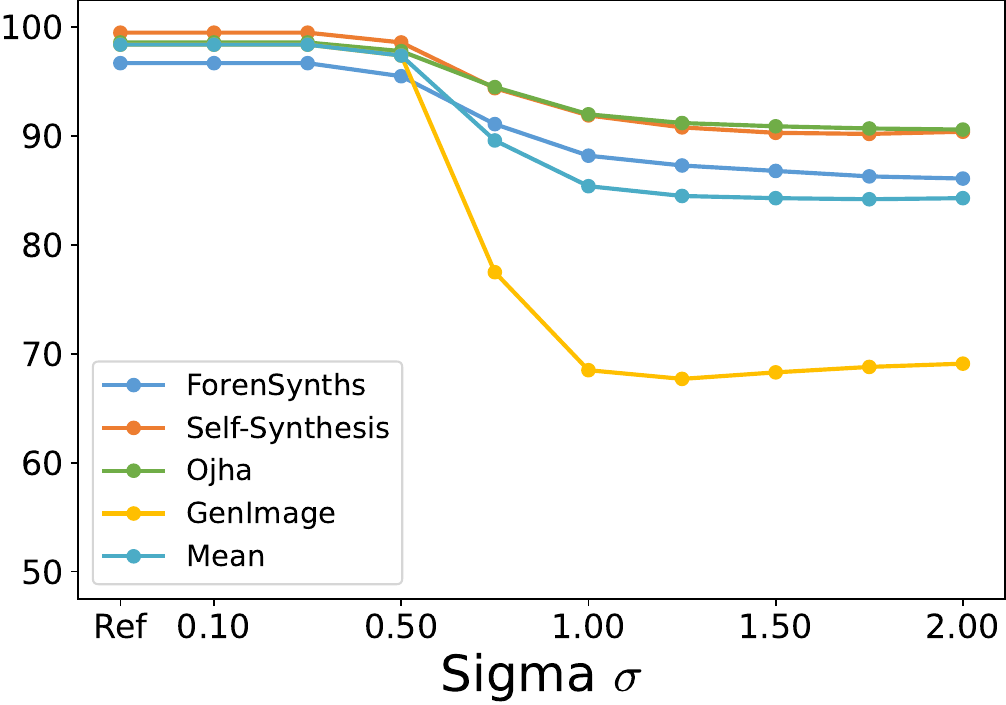}}\\
    \vspace{-3mm}
    \caption{Robustness to Gaussian blur perturbation with different blur sigma $\sigma$ on different test datasets. ``Ref'' corresponds to the detection performance w/o Gaussian blur.}
    \label{fig:gaussian_perturbation_appendix}
\end{figure}

\section{Experimental Details} \label{sec:experimental_details_appendix}

\subsection{Local Correlation Map} \label{sec:details_of_algorithm}

To visualize the differences in local correlations between natural and synthetic images, we introduce sliding windows to traverse the image and calculate the correlation coefficient for pixels within each window. In implementation, we use the Pearson correlation coefficient and set $w$ to $2$ to meet the locality requirement.

\begin{algorithm}
\caption{Calculate Local Correlation Map}
\label{algorithm:local_correlation}
\KwIn{Image $I$ of size $H \times W$, window size $w$}
\KwOut{Correlation map $C$}
\BlankLine
Initialize $C$ as a zero matrix of size $(H-w+1) \times (W-w+1)$\;
\For{$i \gets 0$ \textbf{to} $H - w$}{
    \For{$j \gets 0$ \textbf{to} $W - w$}{
        Extract window $W = I[i:i+w, j:j+w]$\;
        Compute row vector $\mathbf{r}$ as $\mathbf{r}_k = \frac{1}{w} \sum_{m=1}^{w} W_{mk}$ for $k = 1, 2, \ldots, w$\;
        Compute column vector $\mathbf{c}$ as $\mathbf{c}_k = \frac{1}{w} \sum_{n=1}^{w} W_{kn}$ for $k = 1, 2, \ldots, w$\;
        Compute correlation coefficient $\rho$ between $\mathbf{r}$ and $\mathbf{c}$\;
        $C[i, j] \gets \rho$\;
    }
}
\Return $C$\;
\end{algorithm}

\subsection{Details of Figure \ref{fig:ll_ablation}} \label{sec:details_of_table_ll}

This figure provides a horizontal comparison of various artifact features, with a particular emphasis on high-frequency components. In addition to the DWT involved in our pipeline, we also include other commonly-used frequency transforms such as FFT and DCT, along with edge detection operators such as Sobel and Laplace, to facilitate a comprehensive comparison and analysis. Specifically, given an input image $\boldsymbol{X} \in \mathbb{R}^{C \times H \times W}$, the details of these operators are as follows:

\noindent \textbf{Fast Fourier Transform (FFT).} FFT is a method for transforming an image from the spatial domain to the frequency domain. It decomposes the image into sinusoidal waves of different frequencies, and these frequencies can be used to analyze and process the frequency characteristics of the image. In our implementation where FFT is adopted as a high-frequency extractor, the extraction process can be formulated as
\begin{equation}
    \boldsymbol{f}_{\text{FFT}}(\boldsymbol{X}) = \text{IFFT}(\mathcal{B}_{h}^{\text{FFT}}(\text{FFT}(\boldsymbol{X}))),
\end{equation}
where the FFT operator includes both a 2D FFT function on the source image and a zero-frequency shift function to move the zero-frequency component to the center of the spectrum. Then the high-pass filter $\mathcal{B}_{h}^{\text{FFT}}$ is adopted to extract high-frequency components in the form of
\begin{equation}
    \mathcal{B}_h^{\text{FFT}}(f_{i,j}) = 
    \left\{
    \begin{array}{ll}
        0, & \text{if}\ \left|i\right| < \frac{H}{4} \ \text{and} \ \left|j\right| < \frac{W}{4}, \\ 
        f_{i,j}, & \text{otherwise}.
    \end{array}
    \right.
\end{equation}
Subsequently, the corresponding inverse function IFFT is introduced to transform the filtered frequency information back to image space, and we can obtain $\boldsymbol{f}_{\text{FFT}}(\boldsymbol{X}) \in \mathbb{R}^{C \times H \times W}$ as the input artifact feature with FFT.

\noindent \textbf{Discrete Cosine Transform (DCT).} DCT decomposes the image into a combination of cosine functions to analyze the frequency components of the image, which is particularly suited for processing block-based image data. In our implementation, the DCT extractor with the high-frequency filter can be formulated as
\begin{equation}
    \boldsymbol{f}_{\text{DCT}}(\boldsymbol{X}) = \text{IDCT}(\mathcal{B}_{h}^{\text{DCT}}(\text{DCT}(\boldsymbol{X}))),
\end{equation}
where $\mathcal{B}_{h}^{\text{DCT}}$ is the high-frequency filter for the transformed DCT features with a pre-defined threshold $\delta$, which can be formulated as
\begin{equation}
    \mathcal{B}_h^{\text{DCT}}(f_{i,j}) = 
    \left\{
    \begin{array}{ll}
        0, & \text{if}\ i + j < \delta, \\ 
        f_{i,j}, & \text{otherwise}.
    \end{array}
    \right.
\end{equation}
Subsequently, the inverse function IDCT is used to transform the high-frequency component back to image space, and $\boldsymbol{f}_{\text{DCT}}(\boldsymbol{X}) \in \mathbb{R}^{C \times H \times W}$ is regarded as the input artifact feature.

\noindent \textbf{Sobel and Laplace.} Sobel and Laplace operators are both widely used for edge detection to identify regions in the image with significant intensity changes, which are strongly correlated with the high-frequency components. In practice, both operators are implemented through convolution operations, using specific convolution kernels for edge detection. 

The Sobel operator uses two $3 \times 3$ convolution kernels, one for detecting horizontal changes ($G_x$) and the other for detecting vertical changes ($G_y$), that is
\[ 
G_x = \begin{bmatrix}
-1 & 0 & 1 \\
-2 & 0 & 2 \\
-1 & 0 & 1
\end{bmatrix},
\quad
G_y = \begin{bmatrix}
-1 & -2 & -1 \\
0 & 0 & 0 \\
1 & 2 & 1
\end{bmatrix}.
\]

The Laplace operator is a second-order differential operator used to detect edges and details in images. It determines the regions of rapid intensity change by computing the second derivatives of pixel values. The Laplace operator commonly uses two forms of $3 \times 3$ convolution kernels, each with distinct characteristics and applications, that is
\[ 
\begin{bmatrix}
0 & 1 & 0 \\
1 & -4 & 1 \\
0 & 1 & 0
\end{bmatrix}
\quad
\text{or} \quad
\begin{bmatrix}
1 & 1 & 1 \\
1 & -8 & 1 \\
1 & 1 & 1
\end{bmatrix}.
\]

These operators are applied to the image through convolution operations $\boldsymbol{f}_{\text{Sobel}}(\boldsymbol{X}), \boldsymbol{f}_{\text{Laplace}}(\boldsymbol{X}) \in \mathbb{R}^{C \times H \times W}$ for edge detection, which we consider as the input artifact features in our comparison.

\section{Additional Experiments} \label{sec:additional_experiments_appendix}

In this section, we report ACC$_\text{M}$ and AP$_\text{M}$ on all 33 test subsets unless specified, ensuring a comprehensive comparison on both GAN-based and DM-based benchmarks.

\subsection{Hyperparameters} \label{sec:hyperparameter}

We empirically ablate the essential hyperparameters of our proposed data augmentations as shown in Fig.~\ref{fig:hyper_ablation_appendix}, from which we can draw the following observations:
\begin{itemize}[leftmargin=*]
    \item \textbf{Jitter factor $\boldsymbol{\alpha}$}. The jitter factor $\alpha$, employed in ColorJitter (CJ), shows a clear impact on both ACC and AP. As $\alpha$ increases from 0 to 0.8, ACC and AP initially improve, peaking around $\alpha$ = 0.5 for both ACC and AP. Beyond these points, a sharp decline is observed in both metrics, indicating an optimal range for $\alpha$ around 0.4 to 0.6. Excessive jittering deteriorates the detector's performance, highlighting the importance of moderate augmentation.
    \item \textbf{Rotation angle $\boldsymbol{\beta}$.} In RandomRotation (RR), the rotation angle $\beta$ demonstrates a significant influence on the detection performance. Both ACC and AP increase as the rotation angle progresses from $0^{\circ}$ to $60^{\circ}$, stabilizing at high values until $150^{\circ}$. Notably, performance metrics continue to improve when $\beta$ exceeds $150^{\circ}$, reaching their peak at $180^{\circ}$. This suggests that larger rotation angles, including extreme rotations, can enhance the generalization performance by providing diverse training samples.
    \item \textbf{Patch size $\boldsymbol{d}$.} For patch size $d$, the ablation study indicates that smaller patch sizes ($d = 1, 2, 4$) result in lower performance metrics. As $d$ increases to 8, 16, and 32, a marked improvement in both ACC and AP is observed, with optimal performance occurring at $d = 16$. Further increasing the patch size to $d = 64$ results in a slight decline in metrics. This finding suggests that intermediate patch sizes ($d = 16$) provide a balance between detail preservation and context abstraction, thereby enhancing local awareness.
    \item \textbf{Mask ratio $\boldsymbol{R}$.} The mask ratio $R$ in RandomMask (RM) shows an intriguing trend. Both ACC and AP initially fluctuate with increasing $R$, reaching optimal performance around 75\%. Beyond this point, a gradual decline is noted. This implies that a moderate masking ratio, which likely introduces sufficient variation without overwhelming the detector, is beneficial. Over-masking (close to 100\%) can obscure critical information, negatively impacting performance.
\end{itemize}

\begin{figure}[t]
    \centering
    \subfloat[ACC]{\includegraphics[width=.49\columnwidth]{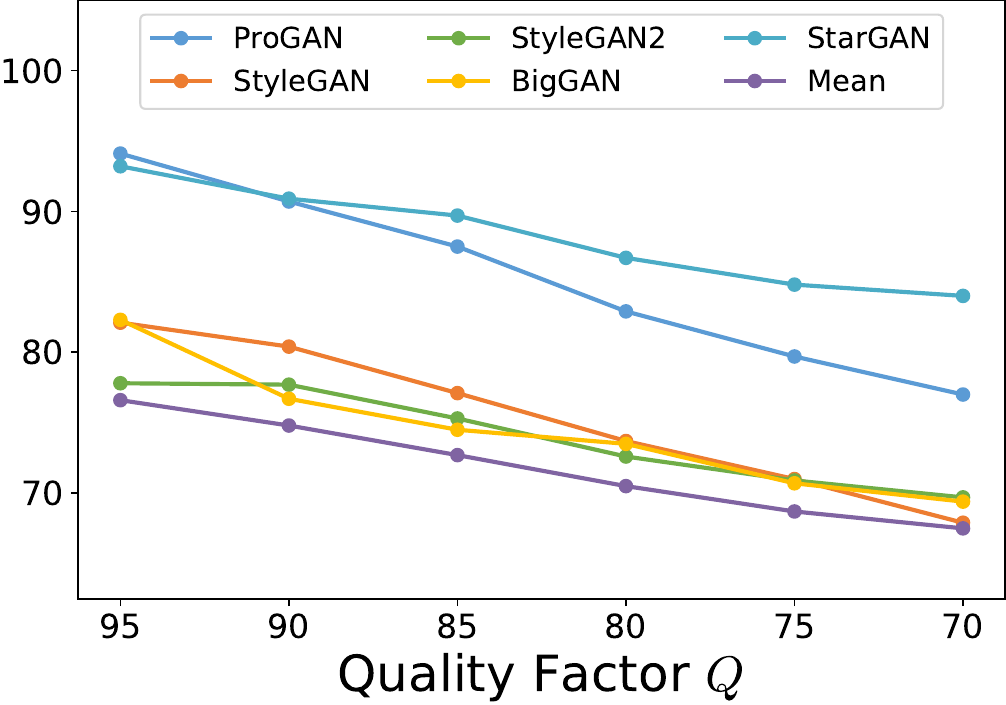}}\hspace{2pt}
    \subfloat[AP]{\includegraphics[width=.49\columnwidth]{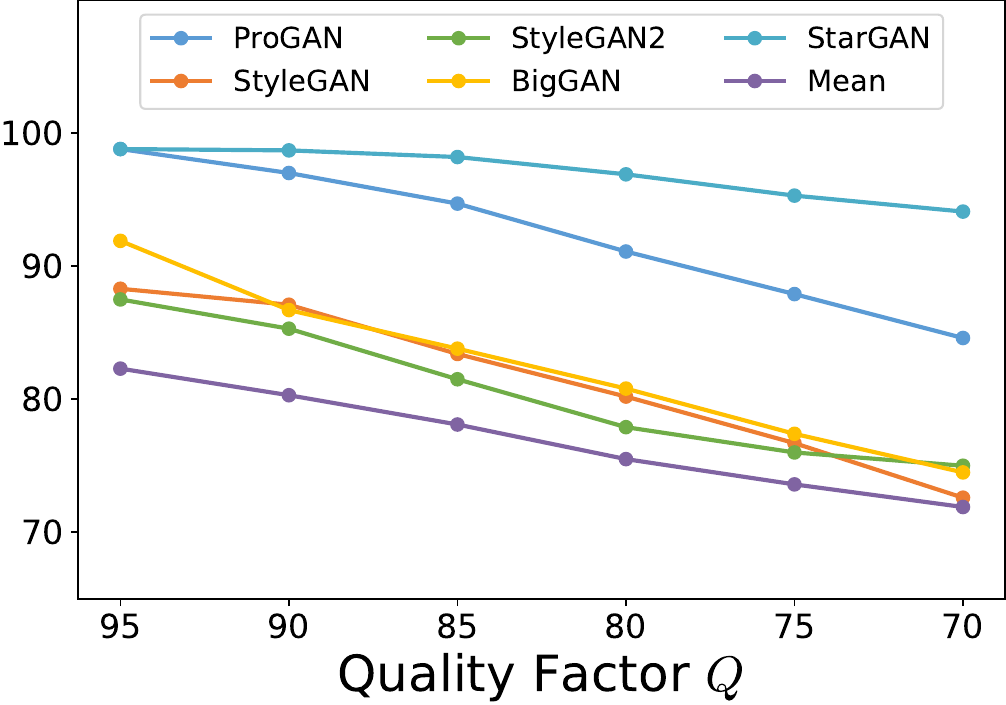}}\\
    \vspace{-3mm}
    \caption{Robustness to JPEG compression perturbation with different quality factors $Q$ on different test datasets. ``Mean" refers to the average results on ForenSynths.}
    \label{fig:jpeg_perturbation_appendix}
\end{figure}
\begin{figure*}[t]
    \centering
    \subfloat[ACC$_\text{M}$ ($d_1=2$)]{\includegraphics[width=.49\columnwidth]{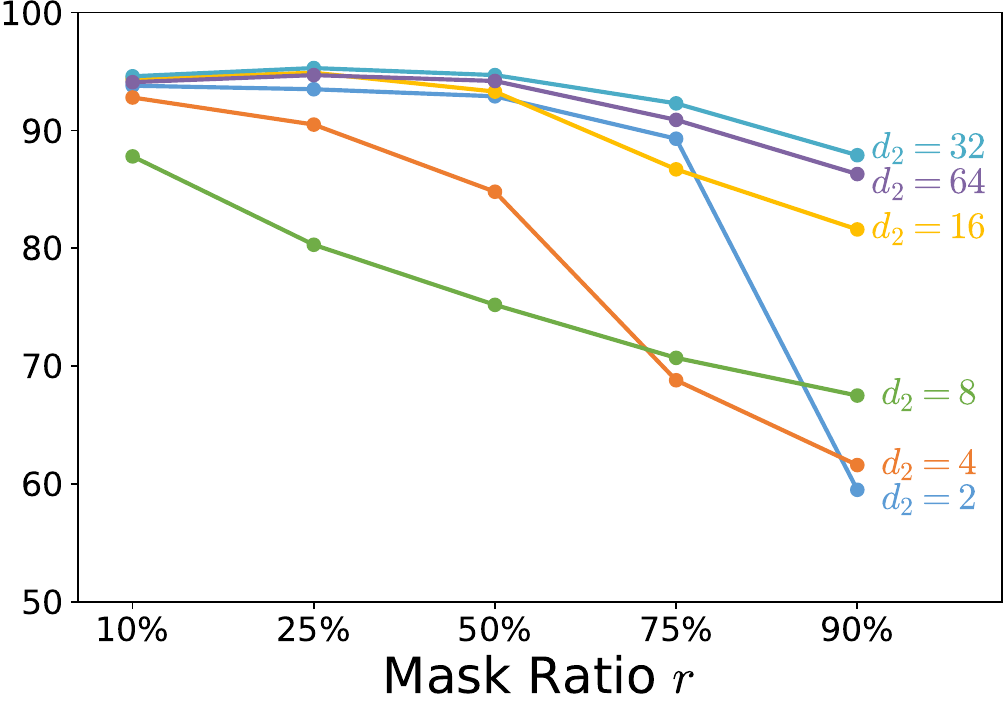}}\hspace{2pt}
    \subfloat[ACC$_\text{M}$ ($d_1=4$)]{\includegraphics[width=.49\columnwidth]{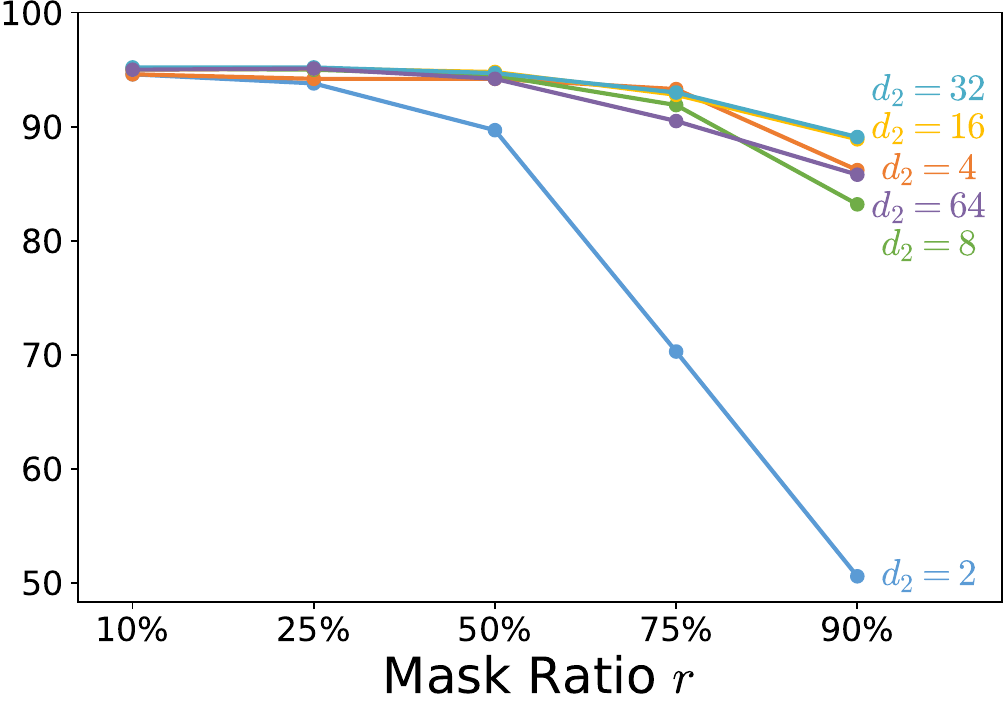}}\hspace{2pt}
    \subfloat[ACC$_\text{M}$ ($d_1=8$)]{\includegraphics[width=.49\columnwidth]{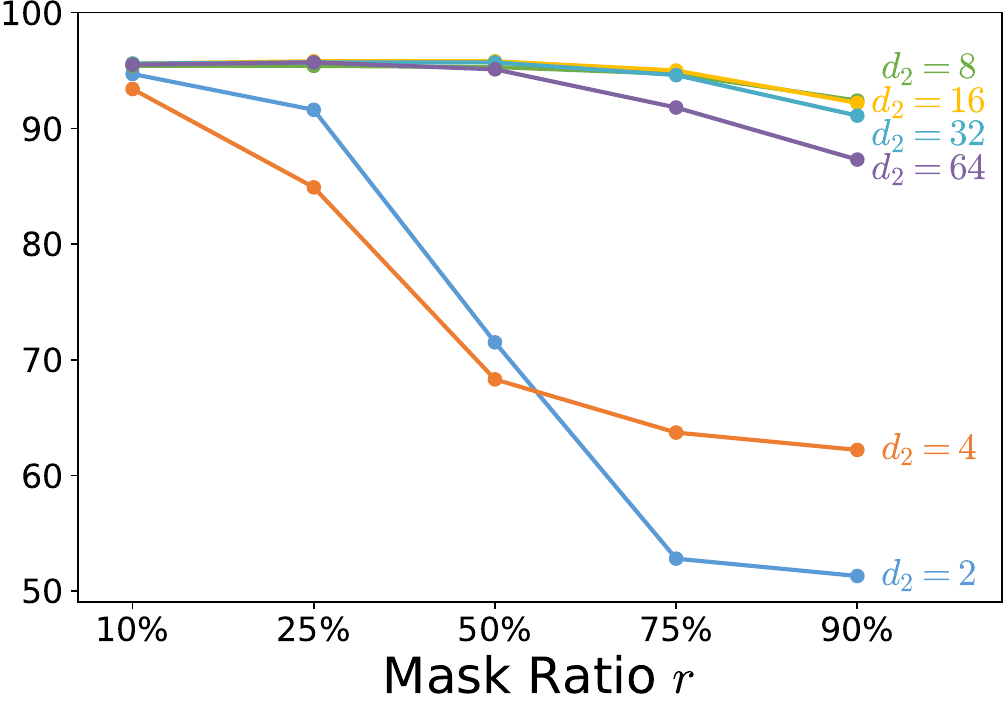}}\hspace{2pt}
    \subfloat[ACC$_\text{M}$ ($d_1=16$)]{\includegraphics[width=.49\columnwidth]{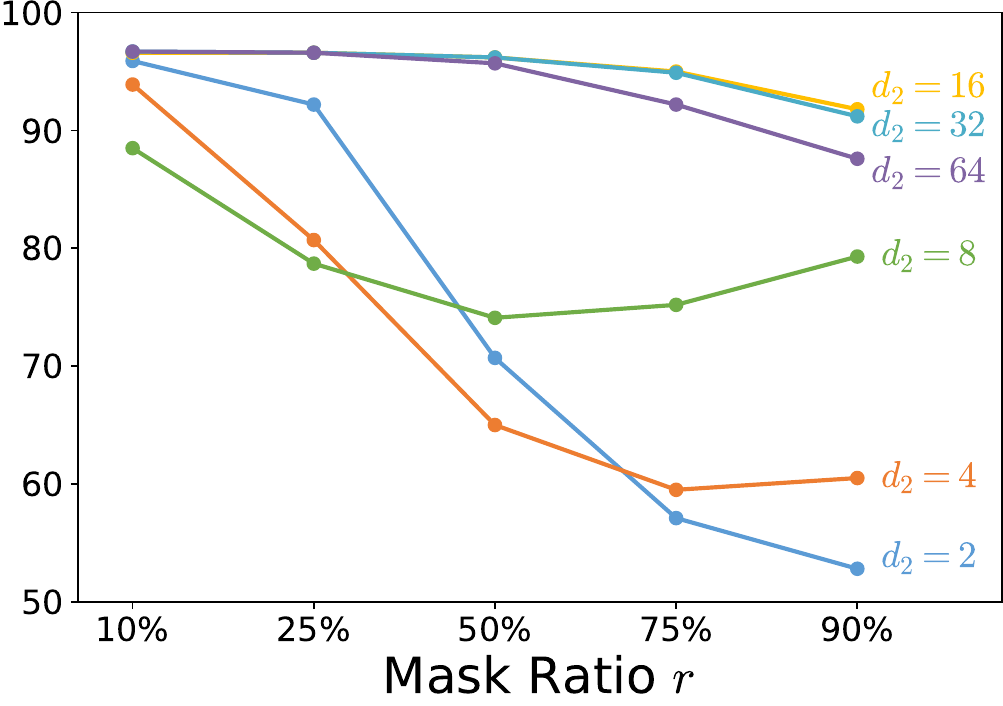}}\\
    \subfloat[AP$_\text{M}$ ($d_1=2$)]{\includegraphics[width=.49\columnwidth]{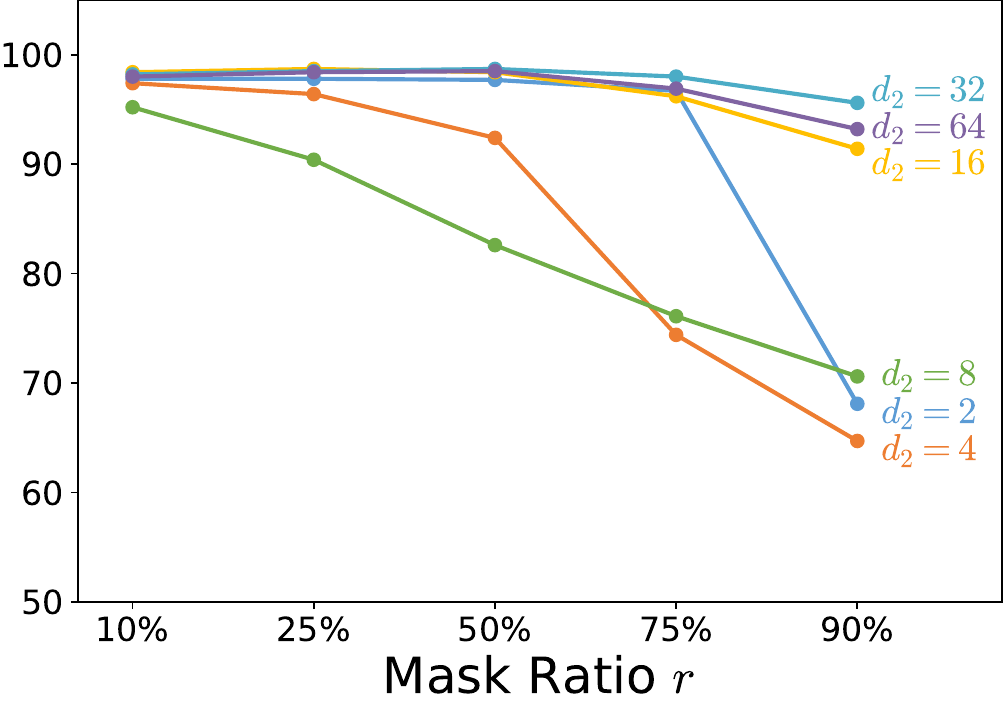}}\hspace{2pt}
    \subfloat[AP$_\text{M}$ ($d_1=4$)]{\includegraphics[width=.49\columnwidth]{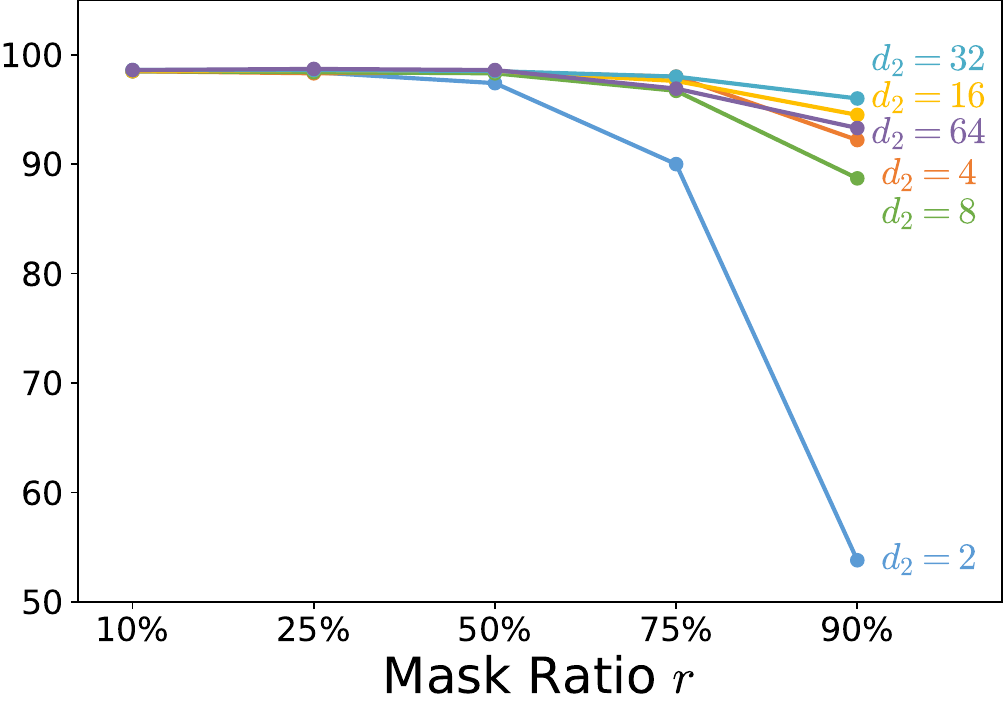}}\hspace{2pt}
    \subfloat[AP$_\text{M}$ ($d_1=8$)]{\includegraphics[width=.49\columnwidth]{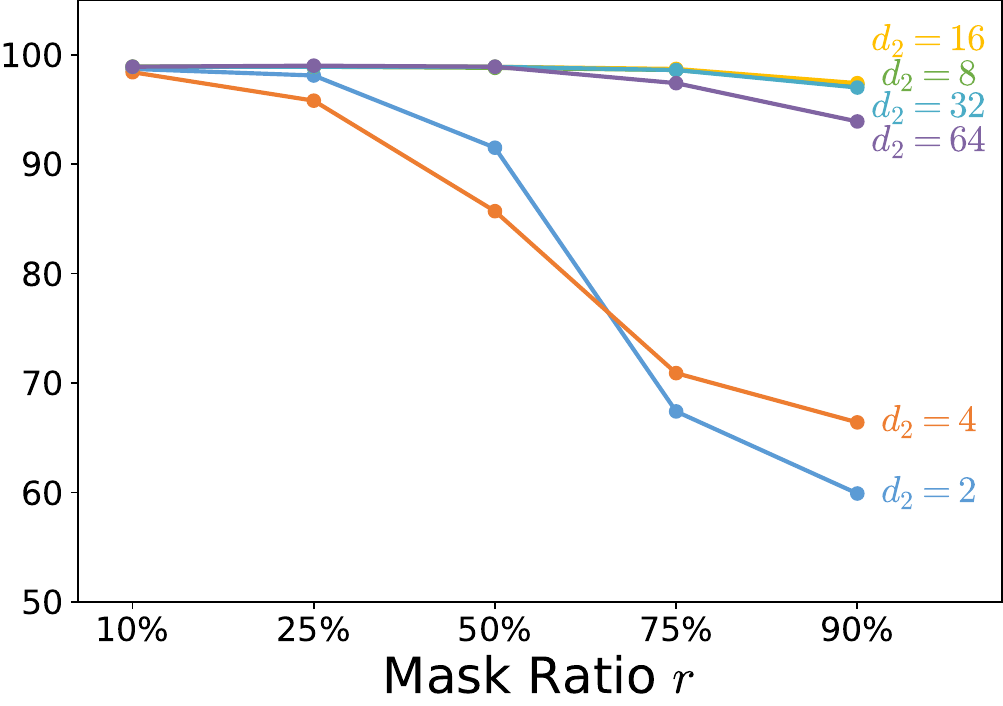}}\hspace{2pt}
    \subfloat[AP$_\text{M}$ ($d_1=16$)]{\includegraphics[width=.49\columnwidth]{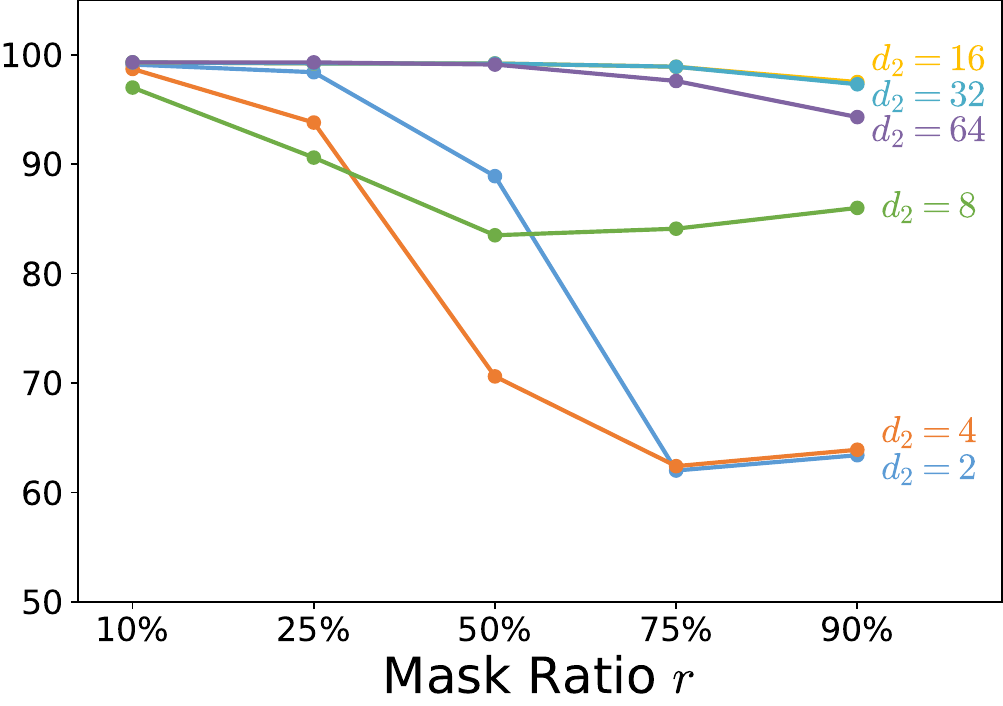}}\\
    \vspace{-3mm}
    \caption{Robustness to random mask perturbation with different mask ratios $r$ and patch size $d_2$, where $d_1$ and $d_2$ refer to the masked patch size at training and testing, respectively.}
    \label{fig:randommask_perturbation_appendix}
\end{figure*}

\subsection{Robustness to Unknown Perturbations}

In real-world scenarios, images shared on public platforms are susceptible to various unknown perturbations, necessitating the evaluation of SID detectors in terms of robustness. Besides the resizing operator discussed Sec.~\ref{sec:ablation_studies}, other plausible operators, such as Gaussian blur, JPEG compression, and random mask, should also be taken into account for a comprehensive assessment.

\noindent \textbf{Gaussian blur.} Gaussian blur can be ubiquitous during image transmission on the Internet. Inspired by this, we simulate the scenario where images are perturbed with varying deviation degrees $\sigma$ of Gaussian blur to evaluate the robustness of the detector in Fig.~\ref{fig:gaussian_perturbation_appendix}. Specifically, we additionally introduce another data augmentation from CNNDect \cite{wang2020cnn}, \textit{i.e.}, RandomGaussianBlur with the standard deviation $\sigma \sim [0.1, 2.0]$ and probability $p=0.5$ at training. This technique is useful in enhancing the robustness to Gaussian blur. It can be observed that the detection performance initially remains relatively stable at low blur levels ($\sigma \leq 0.5$), indicating that our detector is resilient to minimal blurring. As the blur level increases ($0.5 \leq \sigma \leq 1.0$), there is a gradual decline in performance. However, when the blur level reaches $\sigma \geq 1.0$, the detection performance levels off, with results around 85\% ACC$_\text{M}$ and 90\% AP$_\text{M}$. This indicates that our detector maintains considerable robustness against Gaussian blur across varying intensities. Overall, the consistency of our detection pipeline under different levels of Gaussian blur underscores its reliability and suitability for practical applications.

\noindent \textbf{JPEG compression.} Similar to Gaussian blur, we introduce RandomJPEG as an additional data augmentation from CNNDect \cite{wang2020cnn}, with JPEG quality factor $Q \sim [70, 100)$ and probability $p=0.2$ at training. As shown in Fig.~\ref{fig:jpeg_perturbation_appendix}, we observe a significant performance drop in ACC$_\text{M}$ and AP$_\text{M}$. This is because JPEG compression can substantially diminish the high-frequency components of images, which contradicts our adopted high-frequency artifacts from DWT. We will consider addressing this limitation by exploring more robust artifacts in future work.

\begin{table}[t]
\centering
\caption{Statistics of the DiTFake dataset.}
\vspace{-3mm}
\label{table:dataset_statistics}
\resizebox{0.90\hsize}{!}{
\tiny  
\renewcommand{\arraystretch}{1.00}  

\begin{tabular}{c|c|ccc}
\toprule
Generator               &      & Number & Resolution    & Source \\ \midrule
\multirow{2}{*}{Flux}   & Real & 5,000  & $\sim$640×480 & COCO   \\
                        & Fake & 5,000  & 1024×1024     & -      \\ \midrule
\multirow{2}{*}{PixArt} & Real & 5,000  & $\sim$640×480 & COCO   \\
                        & Fake & 5,000  & 1024×1024     & -      \\ \midrule
\multirow{2}{*}{SD3}    & Real & 5,000  & $\sim$640×480 & COCO   \\
                        & Fake & 5,000  & 1024×1024     & -      \\ \bottomrule
\end{tabular}

}

\end{table}
\begin{figure*}[t]
    \centering
    \includegraphics[width=\hsize]{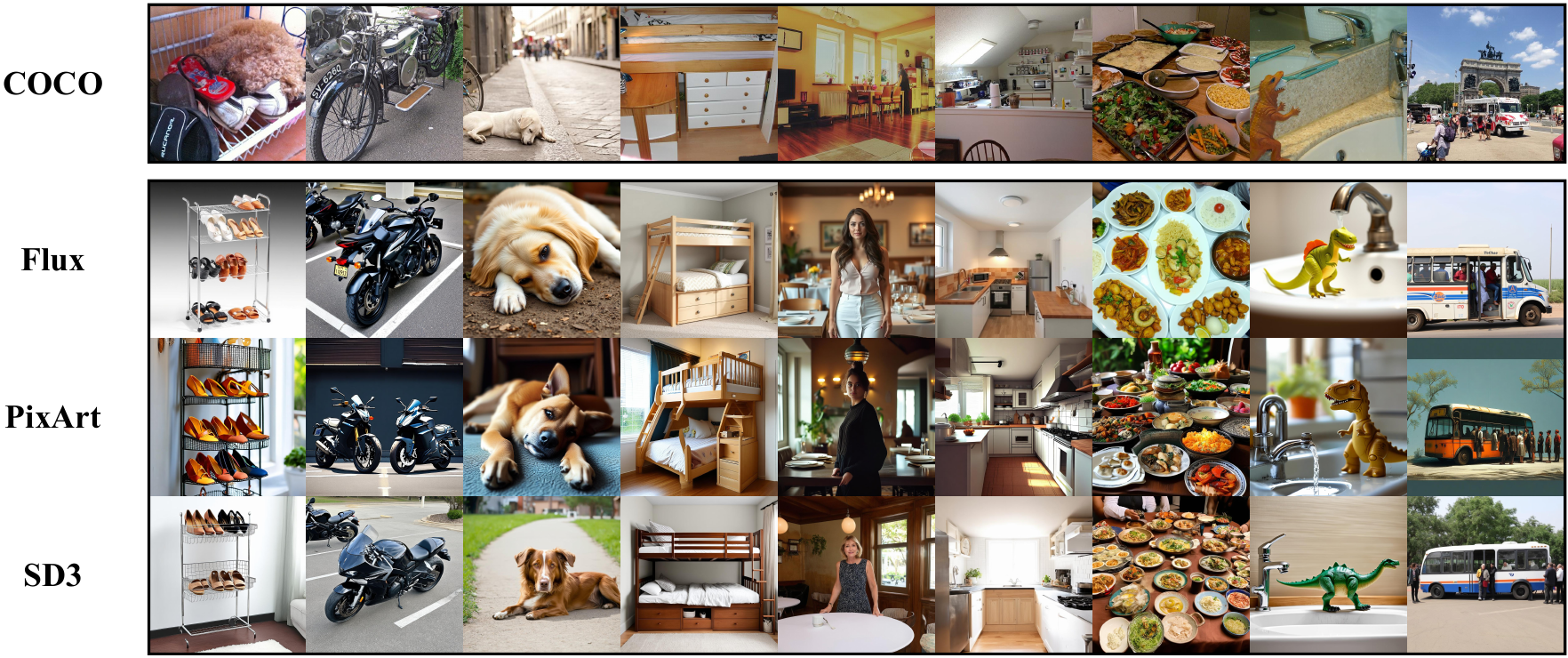}
    \vspace{-6mm}
    \caption{Visualization of the DiTFake dataset, which consists of synthetic images generated from corresponding COCO prompts to mimic real-world scenarios.}
    \label{fig:dataset_appendix}
\end{figure*}

\noindent \textbf{Random mask.} We also simulate the scenario where images are randomly masked with patches, which could visually and semantically degrade the detection performance. Fig.~\ref{fig:randommask_perturbation_appendix} examines the robustness of our detector trained with different patch sizes ($d_1 = 2, 4, 8, 16$) in RandomMask under various mask ratios ($r = $ $10\%,$ $25\%,$ $50\%,$ $75\%,$ $90\%$) and patch sizes ($d_2 = 2, 4, 8, 16, 32, 64$) at inference. We can draw the following conclusions from this figure:
\begin{itemize}[leftmargin=*]
    \item Across all subplots, there is a consistent trend where both ACC$_\text{M}$ and AP$_\text{M}$ decrease as the mask ratio increases. This pattern holds true regardless of the training patch size $d_1$. Higher mask ratios (\textit{e.g.}, $75\%$ and $90\%$) result in more significant performance degradation, indicating the detector's robustness diminishes with increased masked regions.
    
    \item The detector consistently demonstrates robust performance when perturbed with larger patch sizes ($d_2 = 16, 32, 64$) across all $d_1$. This suggests that the detector effectively focuses on the remaining unmasked patches and draws inferences from these local features. The ability to maintain high performance despite significant masking highlights the importance of local awareness in handling such perturbations.
    
    \item In all cases, detection performance deteriorates more significantly as the mask ratio increases, particularly for smaller patches ($d_2 = 2,4$). This can be attributed to the fact that smaller patches are more prone to disrupting the local correlation among pixels, as discussed in Sec.~\ref{sec:data_pre_processing}. These artifacts are inherently introduced by the imaging paradigm of synthetic images, making robust detection challenging with higher mask ratios.
    
    \item Among the four detectors trained with different patch sizes $d_1 = 2, 4, 8, 16$, the detector with $d_1 = 4$ shows the most robustness under random mask perturbation. This comes at the cost of slight performance degradation on unperturbed images compared to $d_1 = 16$ as shown in  Fig.~\ref{fig:hyper_ablation_appendix} (c). The choice of $d_1$ at training should be grounded in specific application scenarios, balancing between detection accuracy and robustness.
\end{itemize}

\section{DiTFake Dataset} \label{sec:ditfake_appendix}

To benchmark our pipeline on latest DiT-based generators, we select three widely-used SOAT DiT-based generators to construct our DiTFake dataset, including Flux\footnote{\url{https://huggingface.co/black-forest-labs/FLUX.1-schnell}}, PixArt\footnote{\url{https://huggingface.co/PixArt-alpha/PixArt-Sigma-XL-2-1024-MS}}, and SD3\footnote{\url{https://huggingface.co/stabilityai/stable-diffusion-3-medium-diffusers}}. We first sample the first 5,000 prompts from the COCO validation set\footnote{\url{https://huggingface.co/datasets/stasstaf/MS-COCO-validation}}, which could generate images that align with real-world scenarios. Then we follow their official sampling configs (\textit{e.g.}, inference steps and guidance scale \cite{ho2022classifier}) from HuggingFace to generate one synthetic image for each prompt, resulting in 5,000 synthetic images for each generator. All synthetic images have a resolution of 1024×1024 and exhibit high visual quality. In terms of real images, we introduce the COCO validation set\footnote{\url{https://huggingface.co/datasets/stasstaf/MS-COCO-validation}} and include the first 15,000 images in our dataset, ensuring they share the same number as synthetic images. Most of these real images have a resolution of approximately 640×480, offering higher resolution and image quality compared to previous datasets as shown in Table~\ref{table:dataset_statistics} and Fig.~\ref{fig:dataset_appendix}.

\section{Limitation} \label{sec:limitation_appendix}
Our pipeline demonstrates that simple image transformations can significantly improve the generalization performance by mitigating training biases in SID. However, the adoption of simple low-level artifacts (\textit{e.g.}, DWT features) proves inferior robustness against plausible unknown perturbations, as these perturbations could undermine discriminative artifacts in low-level space, thereby degrading the detection performance. This performance degradation is prevalent among existing SID pipelines. In addition to investigating other plausible biases during SID training, we hope to explore more robust artifacts to conduct an efficient and robust detector with unbiased training paradigms in future work.

\end{document}